\documentclass[10pt,journal,compsoc]{IEEEtran}
\ifCLASSOPTIONcompsoc
  \usepackage[nocompress]{cite}
\else
  \usepackage{cite}
\fi
\ifCLASSINFOpdf
\else
\fi
\usepackage[ruled]{algorithm2e}
\usepackage{enumerate}
\usepackage{graphicx}
\usepackage{bm}
\usepackage{color}
\usepackage{subfigure}
\usepackage{url}

\makeatletter
\newcommand*\bigcdot{\mathpalette\bigcdot@{.5}}
\newcommand*\bigcdot@[2]{\mathbin{\vcenter{\hbox{\scalebox{#2}{$\m@th#1\bullet$}}}}}
\makeatother

\newtheorem{definition}{Definition}[section]

\begin{document}
%
\title{Representation Learning from Limited\\ Educational Data with Crowdsourced Labels}
%
%
%
%

\author{Wentao Wang,
        Guowei Xu,
        Wenbiao Ding,
        Gale Yan Huang,
        Guoliang Li,
        Jiliang Tang
        and~Zitao Liu
\IEEEcompsocitemizethanks{\IEEEcompsocthanksitem W. Wang and G. Xu make equal contributions to this work.
\IEEEcompsocthanksitem W. Wang and J. Tang are with Data Science and Engineering Lab, Michigan State University, USA. E-mail: \{wangw116, tangjili\}@msu.edu
\IEEEcompsocthanksitem G. Xu, W. Ding, and Z. Liu are with TAL Education Group, China. E-mail: \{xuguowei, dingwenbiao, liuzitao\}@tal.com

\IEEEcompsocthanksitem G. Huang is with TAL Education Group, China. He is also with the Department of Computer Science, Tsinghua University, Beijing, China. E-mail: galehuang@tal.com

\IEEEcompsocthanksitem G. Li is with the Department of Computer Science, Tsinghua
National Laboratory for Information Science and Technology (TNList),
Tsinghua University, China. E-mail: liguoliang@tsinghua.edu.cn

\IEEEcompsocthanksitem Corresponding Author: Zitao Liu

}
\thanks{Manuscript received December 26, 2019; revised August 19, 2020.}}

%
%

\markboth{IEEE TRANSACTIONS ON KNOWLEDGE AND DATA ENGINEERING}%
{Shell \MakeLowercase{\textit{et al.}}: Bare Demo of IEEEtran.cls for Computer Society Journals}
%



\IEEEtitleabstractindextext{%
\begin{abstract}
Representation learning has been proven to play an important role in the unprecedented success of machine learning models in numerous tasks, such as machine translation, face recognition and recommendation. The majority of existing representation learning approaches often require a large number of consistent and noise-free labels. However, due to various reasons such as budget constraints and privacy concerns, labels are very limited in many real-world scenarios. Directly applying standard representation learning approaches on small labeled data sets will easily run into over-fitting problems and lead to sub-optimal solutions. Even worse, in some domains such as education, the limited labels are usually annotated by multiple workers with diverse expertise, which yields noises and inconsistency in such crowdsourcing settings. In this paper, we propose a novel framework which aims to learn effective representations from limited data with crowdsourced labels. Specifically, we design a grouping based deep neural network to learn embeddings from a limited number of training samples and present a Bayesian confidence estimator to capture the inconsistency among crowdsourced labels. Furthermore, to expedite the training process, we develop a hard example selection procedure to adaptively pick up training examples that are misclassified by the model. Extensive experiments conducted on three real-world data sets demonstrate the superiority of our framework on learning representations from limited data with crowdsourced labels, comparing with various state-of-the-art baselines. In addition, we provide a comprehensive analysis on each of the main components of our proposed framework and also introduce the promising results it achieved in our real production to fully understand the proposed framework. To encourage reproducible results, we make our code available online at \url{https://github.com/tal-ai/RECLE}.
\end{abstract}

}

\maketitle

\IEEEdisplaynontitleabstractindextext

%
\IEEEpeerreviewmaketitle

\IEEEraisesectionheading{\section{Introduction}\label{sec:introduction}}

%
%
%
%
\IEEEPARstart{L}{earning} effective representations from data has been proven to be important and helpful in numerous machine learning tasks such as machine translation~\cite{cho-etal-2014-learning,artetxe2018unsupervised}, face recognition~\cite{sun2014deep,tran2017disentangled} and recommendation~\cite{wang2015collaborative,zhang2017joint}, etc., since the performance of machine learning models is heavily dependent on the choice of representation of the input data~\cite{bengio2013representation}. 
A good data representation can provide tremendous flexibilities that allow us to choose fast and simple models~\cite{dong2018feature}. 
Briefly speaking, representation learning aims to automatically learn new data representation from raw features by discovering hidden patterns from data and, hence, reduces the difficulty of useful information extraction when building classifiers or other predictors~\cite{bengio2013representation}. 
In the past few decades, representation learning has been continuously studied by both academia and industry and has attracted more and more attention in recent years with the boom of deep learning techniques~\cite{lecun2015deep}.
For examples, in natural language understanding, words, phrases, sentences are represented as context-aware semantic embeddings to solve many real-world tasks like sentiment analysis~\cite{tang2014learning,dos2014deep} and information retrieval~\cite{huang2013learning,palangi2016deep}. 
In computer vision research, great progress has achieved on image classification tasks after deep convolutional neural networks are introduced to extract reliable high-level features of images~\cite{krizhevsky2012imagenet,radford2015unsupervised}.
In K-12 education, researchers attempted to learn embeddings of students' exercise submissions using deep learning techniques so as to provide useful feedback to help students~\cite{piech2015learning} or predict future performance of students~\cite{wang2017learning}.

Typically, the majority of existing representation learning approaches are often discriminatively trained on massive labeled data~\cite{najafabadi2015deep}.
By adding large-scale consistent and noise-free labeled data into the training process, existing approaches consisting of tens of thousands of parameters and complicated network architectures are able to learn effective embeddings~\cite{simonyan2014very,zeiler2014visualizing,szegedy2015going}.
However, in many real-world scenarios, labeled data is typically insufficient.
For example, many kinds of privacy data of patients are prohibited by applicable laws and financial data are often inaccessible due to confidentiality requirements of companies.
Directly applying standard representation learning approaches on insufficient labeled data sets will easily run into over-fitting problems and lead to sub-optimal solutions~\cite{dundar2015simplicity}.

For lightening the negative impact of a small amount of training data, human efforts can be introduced to acquire labeled data manually and crowdsourcing provides a flexible solution~\cite{rodrigues2018deep}. 
Theoretically, we can obtain an annotated data set as large as we can via crowdsourcing platforms such as Amazon Mechanical Turk\footnote{https://www.mturk.com/}, Figure Eight\footnote{https://www.figure-eight.com/}, etc. 
However, in practice, the number of crowdsourced labels for a given task can still be limited due to a variety of reasons. 
For example, a limited budget prevents us from affording massive labeled data. 
Another example is in some domains such as health care, privacy concerns restrict the labeling process. Data is only accessible by authorized people, which leads to very limited crowdsourced labels.
Although crowdsourcing can alleviate the insufficient labeled data problem to some extent, it brings new challenges. Due to the fact that crowd workers tend to have different levels of expertise~\cite{zheng2016docs}, they may annotate the same object with distinct labels. Therefore, crowdsourced labels can be very inconsistent or noisy~\cite{fan2015icrowd}. The majority of existing representation learning techniques can only work on noise-free labels appropriately rather than crowdsourced labels, so several crowdsourced label processing methods need to be introduced to pre-process the crowdsourced labels \cite{dawid1979maximum,raykar2010learning}. 

The situation is even worse in educational scenarios \cite{xu2020automatic,li2020multimodal,liu2019automatic}. Different from common crowdsourcing related tasks such as medical imaging~\cite{foncubierta2012ground,de2014crowdsourcing} and part-of-speech tagging~\cite{bontcheva2017crowdsourcing,soto2017crowdsourcing}, the label annotation work in educational scenarios usually requires lots of domain knowledge from teaching professionals. For examples, learning an English speech assessment model needs crowdsourced workers to point out wrong phonetic alphabet in each word \cite{liu2020dolphin}. For detecting disfluencies appeared in students' oral presentations, the annotation task asks crowd workers to give disfluency scores to short oral audios. Apparently, it's hard to guarantee that all crowd workers will provide high quality annotations, hence, the label inconsistent problem may be serious. In addition, the label annotation work in educational scenarios often requires much more efforts than annotation tasks in many other domains \cite{chen2019multimodal,huang2020neural}. For example, it may take a crowd worker less than 1 second to annotate an image while the worker has to watch a whole 60-min video before determining the quality of an online class. Therefore, learning effective representations under educational scenarios faces more challenges.

Recent years have witnessed great efforts on learning with limited labeled data~\cite{fei2006one,ravi2016optimization,vinyals2016matching,triantafillou2017few}. Also inferring true labels from inconsistent crowdsourced labels has been studied for decades~\cite{dawid1979maximum,whitehill2009whose,raykar2010learning,rodrigues2014gaussian}. However, research on representation learning with limited and inconsistent crowdsourced labels is rather limited, not to mention studies specific to educational scenarios. Thus, in this paper, we study the problem of representation learning with crowdsourced labels in real-world educational scenarios. In particular, we target on investigating the following three questions: (1) how to take advantage of crowdsourced labels under the limited and inconsistent settings? (2) how to build an unified representation learning framework with crowdsourced labels? and (3) how to make the learning process more effective?

For answering aforementioned research questions, in this work, we present solutions that are applicable to learn effective representations from very limited educational data in order to support various applications. Specifically, in our proposed representation learning framework, we design a grouping based deep neural architecture to generate hundreds of thousands of training instances from only a limited number of labeled data annotated by crowd workers. Furthermore, instead of isolating true label inference from the representation learning process, we use a Bayesian inference to estimate the label confidence and integrate the confidence estimation process into the model learning. For expediting the learning process and improving the quality of the learned representations, an online adaptive hard example selection procedure is integrated into our framework. The main contributions of this paper are summarized below.

\begin{enumerate}
\item An effective representation learning framework is proposed to jointly solve problems of learning embeddings from limited and inconsistent labeled data appeared in many real-world scenarios.
\item A hard example mining strategy is presented to make the training process efficient and sufficient by adaptively selecting hard training groups during each training iteration.
\item Extensive experiments conducted on three educational data sets demonstrate the effectiveness of the proposed framework by comparing with various state-of-the-art baselines.
\end{enumerate}


The rest of this paper is organized as follows. 
We summarize recent research progresses related to our work in Section~\ref{sec:related_work}. 
Section~\ref{sec:problem_state} states the problem we study and lists important notations used in this paper. 
We provide details of our proposed framework in Section~\ref{sec:model}. 
In Section~\ref{sec:experiments}, various experiments are conducted to demonstrate the effectiveness of our framework. 
We conclude this work and discuss future work in Section~\ref{sec:conclusion}.



\section{Related Work}\label{sec:related_work}

In this section, we provide a detailed review of existing methods related to limited labeled data learning, crowdsourced labels learning, and hard example mining.

\subsection{Learning with Limited Labeled Data}

The success of representation learning is typically based on large amounts of labeled data, which is usually unavailable in many domains. 
Therefore, various types of techniques have been developed to enable learning with limited labeled data. 

Motivated by the fact that humans can learn new concepts with very little supervision, few-shot learning aims to learn new concepts from very small amounts of labeled examples. Li et al. learned the useful representation of new object category from a handful of training examples by utilizing a variational Bayesian framework to model object categories \cite{fei2006one}. Vinyals et al. introduced a neural network architecture that combines the idea of metric learning and external memories together for  learning embeddings and achieved improved accuracy performance on various classification tasks~\cite{vinyals2016matching}. Inspired by information retrieval, Triantafillou et al. defined the training objective as optimizing all relative orderings of the points in each training batch and designed an effective model to achieve this object~\cite{triantafillou2017few}. More research work about few-shot learning can be found in~\cite{ravi2016optimization,snell2017prototypical,motiian2017few,duan2017one,oreshkin2018tadam}.

Even though in some domains labeled data are limited, large amounts of unlabeled data are available, which can be utilized to help representation learning. Techniques have been developed to make use of weak supervision such as higher-level abstractions~\cite{takamatsu2012reducing,ratner2016data,ratner2017snorkel}, biased or noisy labels from distant supervision~\cite{liu2017soft,luo2017learning} and data augmentation~\cite{dosovitskiy2015discriminative,ratner2017learning} to learn effective embeddings. 

Another popular machine learning technique used to solve the insufficient training data problem is transfer learning. Briefly, transfer learning allows to utilize knowledge in source domain to improve the performance of learning tasks in target domain. Similar with few-shot learning, the study of transfer learning is also motivated by human behaviors and investigated more than two decades in different names~\cite{pan2009survey}. Recently, transferring knowledge by deep neural networks has attracted more attention due to the impressive achievements of deep learning techniques in many domains~\cite{Tan2018ASO}. More details about transfer learning can be found in the comprehensive surveys~\cite{pan2009survey,weiss2016survey}.

In this work, since we focus on the problem of learning effective embeddings from limited data with crowdsourced labels, weak supervised learning and transfer learning related approaches that requiring extra information or knowledge to train models are not applicable for our problem. 
Moreover, few-shot learning methods aim to learn from an extremely small number of training examples, such as few or even one noise-free examples for unseen categories, while the limited data collected in our problem is associated with crowdsourced labels. Hence, existing few-shot learning methods are not appropriate for our problem.

\subsection{Learning with Crowdsourced Labels}

Crowdsourcing offers a flexible way to get labeled data for model learning. Due to the fact that crowd workers have different levels of expertise, crowdsourced labels are often inconsistent, which may compromise practical applications~\cite{sheng2008get}. Therefore, one of the key problems is to infer true labels from crowdsourced labels~\cite{zheng2017truth}. An EM algorithm is proposed to estimate the error rates when patients answer medical questions with repeated but conflicting responses~\cite{dawid1979maximum}. Inspired by Dawid and Skene~\cite{dawid1979maximum}, Whitehill et al. considered item difficulty for image classification and a score for each annotator is extracted to assess the quality of the annotator~\cite{whitehill2009whose}. Aforementioned approaches infer the true labels independently, which can be sub-optimal solutions for the targeted tasks. Hence, there are increasing attention on combining true label inference with the targeted machine learning tasks. Raykar et al. proposed an EM algorithm to jointly learn the levels of annotators and the regression models~\cite{raykar2010learning}. Likewise, there are efforts to embed the label inference process into other types of models. Rodrigues, Pereira, and Ribeiro generalized Gaussian process classification to consider multiple annotators with diverse expertise~\cite{rodrigues2014gaussian}. Rodrigues et al. studied supervised topic models for classification and regression from crowds~\cite{rodrigues2017learning}. Albarqouni et al. introduced an additional crowdsourcing layer to embed the data aggregation process into convolutional neural network learning~\cite{albarqouni2016aggnet}. Recently, novel techniques have been studied, which do not need iterative EM algorithms to estimate weights of the annotators. Guan et al. modeled information from each annotator and then learned combination weights via back propagation~\cite{guan2018said}.

In addition, the fast development of modern public crowdsourcing platforms provide a convenient way for both individuals and companies to obtain various kinds of crowdsourced labeled data. Accordingly, increasing research interests are raised in managing crowdsourced data. Shan et al. designed a unified crowdsourcing framework to fill missing values for tabular data via taking attribute relationships of given data into consideration~\cite{shan2018t}. Li et al. presented and deployed a novel crowd-powered database system on well-known platforms for addressing various real-world machine-hard problems like data integration and entity collection~\cite{li2018cdb}. More content about crowdsourced data management related topics can be found in recent surveys~\cite{li2016crowdsourced,chai2018crowd,chai2019crowdsourcing}.

The majority of aforementioned learning algorithms have been designed to address the problems of noise and inconsistency in crowdsourced labels and they cannot work as expected when labels are limited. While in this work, we aim to develop algorithms which can jointly solve the challenges from limited and inconsistent labels.

\subsection{Learning with Hard Example Mining}

Hard example mining is a widely used technique in computer vision that aims to find hard training examples over an overwhelming number of easy ones. An automatic selection procedure can make the model training process more effective and efficient. Unsurprisingly, this is not a new challenge and a standard solution, originally called bootstrapping (and now often called hard negative mining), has been studied for at least 20 years~\cite{sung1996learning}. Felzenszwalb et al. demonstrated that bootstrapping for SVMs converges to the global optimal solution. The corresponding algorithm is referred to as hard negative mining~\cite{felzenszwalb2009object}, which is often used in object detection~\cite{girshick2014rich,he2015spatial}. Shrivastava et al. presented an online hard example mining algorithm for training region-based object detectors. This algorithm dynamically samples training examples according to a non-uniform, non-stationary loss-aware distribution~\cite{shrivastava2016training}. Yuan et al. cascaded three deep models for learning embeddings such that the following model can only focus on the hard examples from previous one~\cite{yuan2017hard}.

In this paper, we propose a new hard example mining strategy for automatically creating hard training groups in order to make the model training process more effective.

\begin{table}[htb]
    \caption{Important notations.}
    \label{tab:notations}
    \centering
    \begin{tabular}{|l|p{5.1cm}|}
    \hline
         Symbol & Definition or Description \\
         \hline
         \hline
         $\mathcal{X}$, $\mathcal{V}$ & the training data set and the validation data set \\
         $\mathcal{X}^{+}$, $\mathcal{X}^{-}$, $\mathcal{V}^{+}$, $\mathcal{V}^{-}$ & the collections of positive examples and negative examples in $\mathcal{X}$ and $\mathcal{V}$ \\
         $\mathcal{G}$ & the collection of training groups generated from $\mathcal{X}$ \\
         \hline
         $\mathbf{x}_i$, $\mathbf{x}^{+}_i$, $\mathbf{x}^{-}_i$, $\mathbf{v}_i$, $\mathbf{v}^{+}_i$, $\mathbf{v}^{-}_i$ & the $i$-th example in $\mathcal{X}$, $\mathcal{X}^{+}$, $\mathcal{X}^{-}$, $\mathcal{V}$, $\mathcal{V}^{+}$ and $\mathcal{V}^{-}$ \\
         $\mathbf{y}_i$, $\mathbf{y}^{+}_i$, $\mathbf{y}^{-}_i$, $\mathbf{u}_i$, $\mathbf{u}^{+}_i$, $\mathbf{u}^{-}_i$ & the crowdsourced label for $\mathbf{x}_i$, $\mathbf{x}^{+}_i$, $\mathbf{x}^{-}_i$, $\mathbf{v}_i$, $\mathbf{v}^{+}_i$ and $\mathbf{v}^{-}_i$ \\
         $d$ & the number of crowd workers\\
         $y_{i, l}$& the label annotated by the $l$-th crowd worker in $\mathbf{y}_i$ \\
         $\mathbf{g}_j$ & the $j$-th training group in $\mathcal{G}$ \\
         $\mathbf{z}_i$, $\mathbf{z}^{+}_i$, $\mathbf{z}^{-}_i$, $\mathbf{w}_i$, $\mathbf{w}^{+}_i$, $\mathbf{w}^{-}_i$ & the learned representation of $\mathbf{x}_i$, $\mathbf{x}^{+}_i$, $\mathbf{x}^{-}_i$, $\mathbf{v}_i$, $\mathbf{v}^{+}_i$ and $\mathbf{v}^{-}_i$ \\
         $M$, $N$, $P$, $Q$ & total number of examples in $\mathcal{X}^{+}$, $\mathcal{X}^{-}$, $\mathcal{V}^{+}$ and $\mathcal{V}^{-}$ \\
         $K$ & total number of groups in $\mathcal{G}$ \\
         $S$ & total number of examples in $\mathbf{g}_j, j \! \in \![1, K]$ \\
         $a^j_1, a^j_2, \dots, a^j_S$ & indices of selected examples in the training group $\mathbf{g}_j$ \\
         \hline
         $T$ & total number of learning iterations \\
         $(t)$ & the index of current learning iteration \\
         $r(\bigcdot, \bigcdot)$ & cosine relevance between two learned embeddings \\
         $\bm{\Omega}$ & model parameters \\
         $\mathcal{L}(\bm{\Omega})$ & the loss function for model training \\
         \hline
    \end{tabular}
    \label{tab:my_label}
\end{table}

\section{Problem Statement and Notations}\label{sec:problem_state}

Given a small data set with inconsistent crowdsourced labels, we are trying to jointly address the challenges from limited and inconsistent crowdsourced labels and discover a feasible way to learn effective embeddings. 
Without loss of generality, for any training example $\mathbf{x}_i$, we assume its corresponding crowdsourced label $\mathbf{y}_i$ is annotated by $d$ crowd workers with binary value 0 or 1, i.e., $\mathbf{y}_i = [y_{i, 1}, y_{i, 2}, \dots, y_{i, d}]$, where $y_{i, l} \in \{0, 1\}$, 1 indicates the positive label and 0 indicates the negative label, and $l = [1, 2, \dots, d]$. 
In addition, we use $(\bigcdot)^{+}$ and $(\bigcdot)^{-}$ to indicate positive and negative examples, respectively. 
Hence, the problem we study in this paper can be formally defined as below.
\begin{definition}[Learning from Crowdsourcing Data]
Given a small amount of labeled data examples set $\mathcal{X}$ annotated by $d$ crowd workers, our goal is to train a learning model that can effectively generate representative embeddings from high-dimensional raw features.
\end{definition}
For the sake of convenience, we summarize the important notations used throughout the paper in Table~\ref{tab:notations}.

In practice, even though two data examples are both identified as positive examples, the confidence of their ``positiveness" might be different. 
For instance, assuming $\mathbf{x}_i^{+}$ and $\mathbf{x}_j^{+}$ are two positive examples whose 5-person crowdsourced labels are $[1,1,1,1,1]$ and $[1,0,1,0,1]$, respectively. 
Intuitively, comparing with data example $\mathbf{x}_j^{+}$, $\mathbf{x}_i^{+}$ is more confident to be recognized as a positive example, since all crowd workers annotate $\mathbf{x}_i^{+}$ to 1.
Therefore, we cannot ignore this difference between crowdsourced labels and should involve such inconsistency in the model training process. 
Specifically, in this work, we quantify such inconsistency as the confidence of the crowdsourced labels. 
\begin{definition}[Crowdsourced Label Confidence]\label{def:confidence}
The likelihood that the corresponding data example is a positive example.
\end{definition}
By taking the confidences of crowdsourced labels into account, our model can be trained more powerful for learning meaningful and representative feature embeddings.

Furthermore, to speed up model training process, we present a hard example selection strategy to pick up partial training examples. 
Different from the traditional definition of hard examples, we define hard examples used in this paper as follows.
\begin{definition}[Hard Example]\label{def:hard_example}
A data example $\mathbf{x}_i$ in training set $\mathcal{X}$ is a hard example if $\mathbf{x}_i$ is the most similar example of example $\mathbf{v}_i$ from validation set $\mathcal{V}$, where $\mathbf{v}_i$ is misclassified by the current version of the trained model.
\end{definition}
Note that different distance metrics may produce different ``most similar example" for a mis-predicted example $\mathbf{v}_i$.
In our model, we use cosine similarity as the default distance metric and also investigate the impacts of different distance metrics in Section~\ref{sec:experiments}.

Next, we will illustrate the overview of our proposed learning model and provide more details about how the confidence scores of crowdsourced labels are estimated and hard examples are selected in our framework.

\section{The Proposed Framework}\label{sec:model}

In this section, we introduce our model which is able to learn effective embeddings from limited data examples with crowdsourced labels. 
More specifically, instead of training the discriminative representation model from a small amount of labeled data directly, we present a new idea that creates tremendous training groups as the input of our model for training (Section~\ref{subsec:grouping}). 
In order to deal with the inconsistency issue of crowdsourced labels, a crowdsourced label confidence estimation approach is integrated into our model to guide the model training (Section~\ref{subsec:bayesian}). 
Moreover, in each training iteration, our model adopts an adaptive hard example selection strategy for make the training process more efficiently and sufficiently (Section~\ref{subsec:adaptive}). 

\subsection{Grouping Based Deep Neural Architecture}\label{subsec:grouping}

\subsubsection{Grouping Strategy}

In our educational practice, due to many practical reasons as we mentioned before, the number of annotated labels coming from the crowdsourced workers is very limited.
The scarcity problem of annotated labels may easily lead to the over-fitting problems for many existing discriminative representation models. 
Intuitively, this issue can be addressed when lots of labeled examples are available for model training. 
Inspired by this intuition, we develop a grouping based deep architecture that can re-assemble and transform limited labeled examples into many training groups to feed the discriminative representation model. 
In our proposed grouping strategy, both positive and negative examples are ensured to be included in each group, and we maximize the conditional likelihood of one positive example given another positive example while minimize the conditional likelihood of one positive example given several negative examples. 
Different from traditional metric learning approaches that focus on learning distances between pairs~\cite{yang2006distance,kulis2012metric}, our approach aims to generate a more difficult scenario that considering not only the distances between positive examples but distances between negative examples, which can enforce the positive examples are close to each other in the learned embedding space while far away from negative examples during the model training phase.

Specifically, based on a limited training data set $\mathcal{X}$, a training group $\mathbf{g}_j$, $j=1, 2, \cdots, K$, can be obtained by the following two steps.

\begin{enumerate}[Step 1.]
\item Selecting 2 positive examples $\mathbf{x}_{a_{1}^j}^{+}$ and $\mathbf{x}_{a_{2}^j}^{+}$ from $\mathcal{X}^{+}$;
\item Selecting $S - 2$ negative examples $\mathbf{x}_{a_{3}^j}^{-}, \mathbf{x}_{a_{4}^j}^{-}, \dots, \mathbf{x}_{a_{S}^j}^{-}$ from $\mathcal{X}^{-}$.
\end{enumerate}
Therefore, each training group $\mathbf{g}_j$ is defined as $\mathbf{g}_j = <\mathbf{x}_{a_{1}^j}^{+}, \mathbf{x}_{a_{2}^j}^{+}, \mathbf{x}_{a_{3}^j}^{-}, \mathbf{x}_{a_{4}^j}^{-}, \dots, \mathbf{x}_{a_{S}^j}^{-}>$, and the entire training group collection $\mathcal{G}$, i.e., $\mathcal{G} = \{\mathbf{g}_1, \mathbf{g}_2, \dots, \mathbf{g}_K\}$. 

\begin{figure}[!t]
\centering
\includegraphics[width=3.5in]{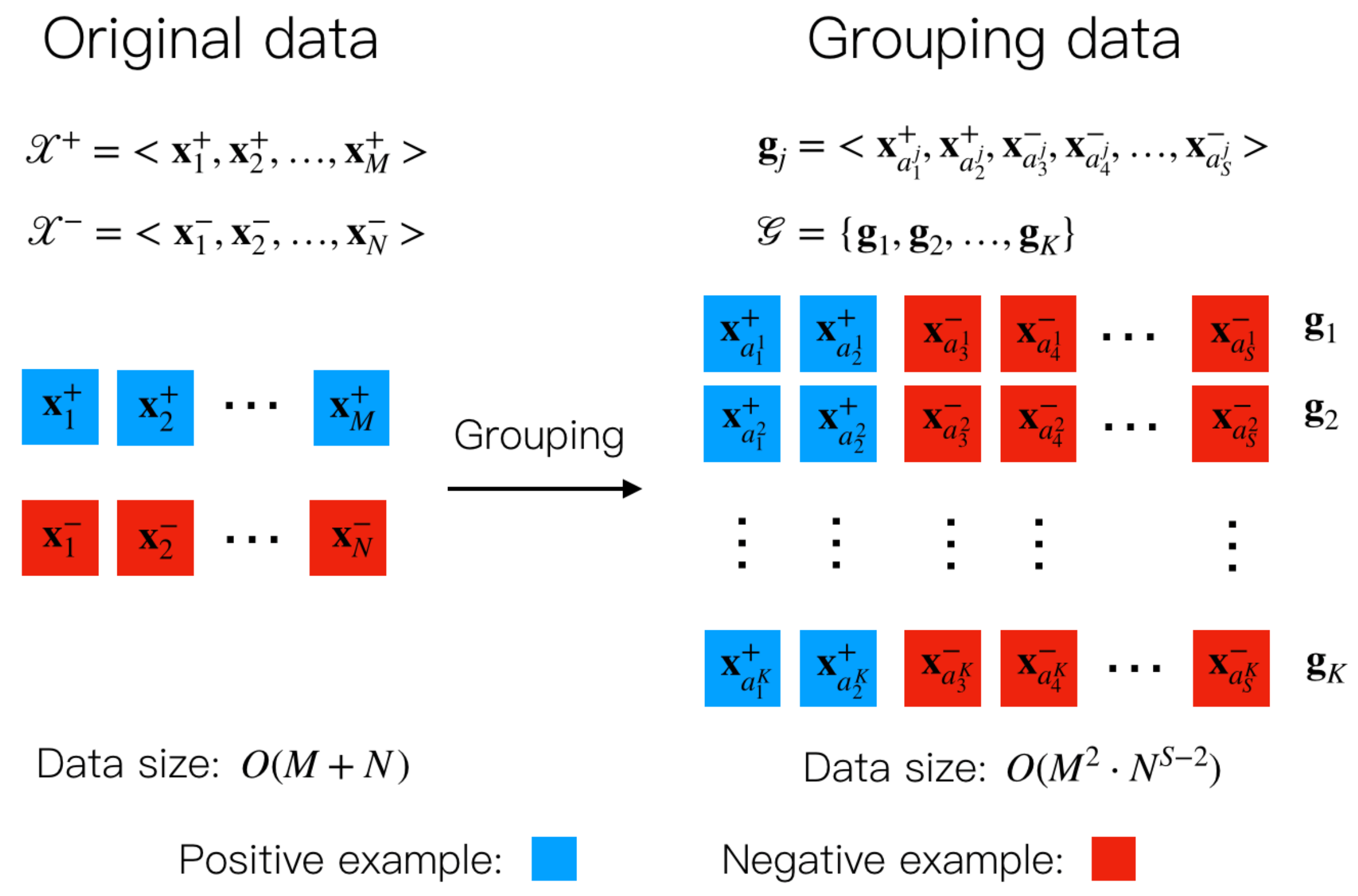}
\caption{An illustration for the proposed grouping strategy.}
\label{fig:grouping}
\end{figure}

Theoretically, by using the above grouping strategy, we can create $O(M^2N^{S-2})$ groups for model training when given $M$ positive training examples and $N$ negative training examples. Figure~\ref{fig:grouping} provides a legible illustration for the aforementioned grouping strategy. 

Note that several selection strategies can be applied in here for picking up positive and negative examples separately to create training groups and the simplest one is random selection.
However, the randomly formed training groups cannot provide extra support for model training. In order to facilitate the model training process and obtain effective embeddings, we propose a hard example mining based selection approach, which can be regarded as a more precise and concise way to choose positive and negative examples for creating training groups. 
We will discuss this approach in Section~\ref{subsec:adaptive}.

After the aforementioned grouping procedure, we treat each group $\mathbf{g}_j$ as a training example and feed $\mathbf{g}_j$ into a typical deep neural network (DNN) for learning robust embeddings. 
The inputs to the DNN are raw features extracted from data examples and the outputs of the DNN are low-dimensional embedded feature vectors. 
Inside the DNN, we use the multi-layer fully-connected non-linear projections to learn representative embeddings.

\begin{figure*}[!bpht]
\centering
\includegraphics[width=7.1in]{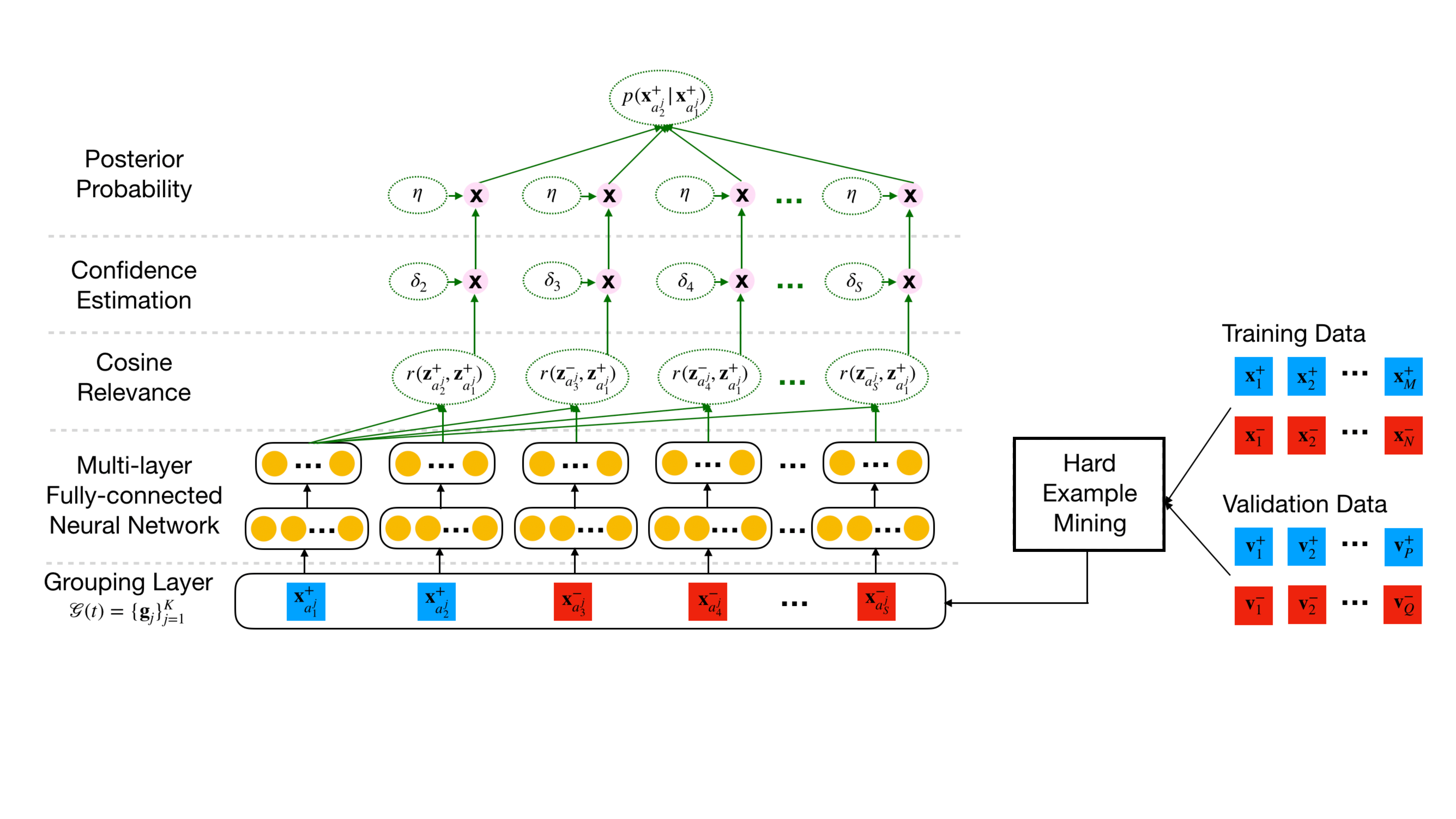}
\vspace{-2cm}
\caption{An overview of the proposed representation learning framework.}
\label{fig:architecture}
\end{figure*}

\subsubsection{Model Learning} 
Inspired by the discriminative training approaches widely used in information retrieval~\cite{huang2013learning,palangi2016deep} and natural language processing~\cite{dos2014deep}, we present a supervised training approach to learn parameters of our model by maximizing the conditional likelihood of retrieving positive example $\mathbf{x}_{a_2^j}^{+}$ given positive example $\mathbf{x}_{a_1^j}^{+}$ from group $\mathbf{g}_j$, $j=1, 2, \cdots, K$.

Formally, in our representation learning model, we compute the posterior probability of $\mathbf{x}_{a_2^j}^{+}$ in group $\mathbf{g}_j$ given $\mathbf{x}_{a_1^j}^{+}$ from the cosine relevance score between them through a softmax function, i.e., 

$$
p \big(\mathbf{x}_{a_2^j}^{+}|\mathbf{x}_{a_1^j}^{+} \big) = \frac{\exp \big(\eta \cdot r (\mathbf{z}_{a_2^j}^{+}, \mathbf{z}_{a_1^j}^{+})\big)}{\exp \big(\eta \cdot r(\mathbf{z}_{a_2^j}^{+}, \mathbf{z}_{a_1^j}^{+})\big) + \sum\limits_{i = 3}^S \exp \big(\eta \cdot r(\mathbf{z}_{a_i^j}^{-}, \mathbf{z}_{a_1^j}^{+})\big)}
$$

\noindent where $\mathbf{z}_{a_1^j}^{+}$ and $\mathbf{z}_{a_2^j}^{+}$ are learned representations of $\mathbf{x}_{a_1^j}^{+}$ and $\mathbf{x}_{a_2^j}^{+}$, respectively, $r(\bigcdot, \bigcdot)$ is the cosine relevance score and $\eta$ is a smoothing hyper-parameter in the softmax function, which is set empirically on a held-out data set in our experiments.

To maximize the posterior, we would like to maximize the relevance between two positive embeddings $\mathbf{z}_{a_1^j}^{+}$ and $\mathbf{z}_{a_2^j}^{+}$ and, in the meanwhile, minimize the relevance between the positive embedding $\mathbf{z}_{a_1^j}^{+}$ and all the other negative embeddings, i.e., $\mathbf{z}_{a_3^j}^{-}, \mathbf{z}_{a_4^j}^{-}, \dots \mathbf{z}_{a_S^j}^{-}$. As distance is proportional to the inverse of relevance, similar data examples are pull closer while dissimilar examples are pushed away in the embedding space.

Hence, given a collection of groups $\mathcal{G}$, we optimize parameters of the DNN by maximizing the sum of log conditional likelihood of finding a positive example $\mathbf{x}_{a_2^j}^{+}$ given the paired positive example $\mathbf{x}_{a_1^j}^{+}$ from group $\mathbf{g}_j$, i.e.,

$$\mathcal{L}(\bm{\Omega}) = - \sum_{j = 1}^{K} \log p \big(\mathbf{x}_{a_2^j}^{+}|\mathbf{x}_{a_1^j}^{+} \big)$$ 

\noindent where $\bm{\Omega}$ is the parameter sets of the DNN. Since $\mathcal{L}(\bm{\Omega})$ is differentiable with respect to $\bm{\Omega}$, we use gradient based optimization approach to train the DNN.

\subsection{Bayesian Confidence Estimation}\label{subsec:bayesian}

Because the inconsistency problem is intrinsic for crowdsourced labels, in this paper, we use the \emph{Crowdsourced Label Confidence} (Definition~\ref{def:confidence}) to model the inconsistency of crowdsourced labels. 
Let $\delta_i$ be the confidence of any data example $\mathbf{x}_i$. 
A common solution is to treat the confidence $\delta_i$ as a random variable that follows the Bernoulli distribution and infer $\delta_i$ by utilizing maximum likelihood estimation (MLE) as follows:

\begin{equation}\label{equ:bernoulli}
\hat{\delta}_i^{Bernoulli} = \sum_{j = 1}^d y_{i, j} / d.
\end{equation}

However, in many real-world scenarios, we are not able to afford too many crowd workers to annotate the same example simultaneously, i.e., $d$ is relatively small. 
This situation leads to the inferior performance in the MLE approach shown in Eq.(\ref{equ:bernoulli}). 
To address this problem, similar to \cite{raykar2010learning}, we assign a Beta prior to $\delta_i$, i.e., $\delta_i \sim Beta(\alpha, \beta)$. 
Hence, the posterior estimation of the crowdsourced label confidence $\delta_i$ is 

$$
\hat{\delta}_i = \frac{\alpha + \sum_{j=1}^{d} y_{i, j}}{\alpha + \beta + d}.
$$

After integrating the confidence estimation part into our representation learning model, we are able to obtain the confidence weighted conditional probability defined as:
\begin{equation}\label{equ:weight_object}
\resizebox{1.0\hsize}{!}{$p \big(\mathbf{x}_{a_2^j}^{+}|\mathbf{x}_{a_1^j}^{+} \big) = \frac{\exp \big(\eta \cdot \delta_2 \cdot r (\mathbf{z}_{a_2^j}^{+}, \mathbf{z}_{a_1^j}^{+})\big)}{\exp \big(\eta \cdot \delta_2 \cdot r(\mathbf{z}_{a_2^j}^{+}, \mathbf{z}_{a_1^j}^{+})\big) + \sum\limits_{i = 3}^S \exp \big(\eta \cdot \delta_i \cdot r(\mathbf{z}_{a_i^j}^{-}, \mathbf{z}_{a_1^j}^{+})\big)}$.}
\end{equation}
Accordingly we adjust the objective function by using the confidence weighted conditional probability, i.e., Eq.(\ref{equ:weight_object}). 

By taking the inconsistency of crowdsourced labels into consideration, the difference between different data examples could be captured and, hence, the learned embeddings are more meaningful and representative. 
We will demonstrate the importance and necessity of considering the crowdsourced labeling inconsistency in Section 5.

\subsection{Adaptive Hard Example Selection}\label{subsec:adaptive}

After transforming the DNN learning procedure from instance level to group level, we are able to get quadratic or cubic training sample size compared to the original data set. 
On the one hand, we provide the DNN sufficient training data and avoid the over-fitting problem. On the other hand, the training process becomes extremely long due to the fact that a complete combination of groups may be incredibly large. 
If we feed all the groups into the DNN, the training process is not efficient and does not guarantee an optimal performance. 
Therefore, we need a method to sample good training examples from massive number of groups.

As the learning iterations proceed, more and more characteristics of data are captured by the learning model and most training groups become easy examples for the model. 
These easy examples are not effective for training and should not appear in the training data set any more. 
Therefore, we design a hard example selection approach to adaptively select hard examples and augment the training data set by the following four steps:
\begin{enumerate}[Step 1.]
\item During each iteration $t$, $t=1, 2, \cdots, T$, we evaluate the quality of the learned embeddings by applying $\bm{\Omega}(t)$ on the validation set $\mathcal{V}$. For each validation example $\mathbf{v}_i$ at iteration $t$, we extract its corresponding embedding $\mathbf{u}_i$ from $\bm{\Omega}(t)$ for the predictive performance evaluation. Then we collect all mis-predicted examples into a set $\mathcal{V}_{miss}(t)$;
\item For each $\mathbf{v}'_j \in \mathcal{V}_{miss}(t)$, we select example $\mathbf{x}'_j$ from original training set $\mathcal{X}$ that is most similar to $\mathbf{v}'_j$. Here we use the cosine function to measure the similarity (the choice of distance functions is discussed in Section~\ref{subsec:distance}). The collection of these examples selected from $\mathcal{X}$ are referred to as hard examples (Definition~\ref{def:hard_example}) at iteration $t$, i.e., $\mathcal{X}_{hard}(t)$;
\item Similar to the grouping approach discussed in Section~\ref{subsec:grouping}, we obtain the hard training groups $\mathcal{G}_{hard}(t)$ by creating a complete list of combinations out of $\mathcal{X}_{hard}(t)$;
\item The training groups set used at the $(t + 1)$-th iteration is $\mathcal{G}(t + 1) = \left\{\mathcal{G}_{base}, \mathcal{G}_{hard}(t) \right\}$, where $\mathcal{G}_{base}$ represent the initial groups from original training set $\mathcal{X}$ by random selection before the model training process starts.
\end{enumerate}

The above adaptive learning process is repeated until predictive performance converges or the maximum number of iterations is reached. Compared to traditional DNN training process, where training data remains the same, our adaptive hard example selection strategy ensures that the neural network can obtain different training groups at each iteration. More importantly, the DNN is fed with hard groups and able to quickly adjust itself to make amendments for mistakes made in the previous iterations, which can enforce the learned representations to be more robust and effective. Please note that, in Step 4, we use the combination of $\mathcal{G}_{based}$ and $\mathcal{G}_{hard}(t)$ instead of only $\mathcal{G}_{hard}(t)$ as the selected training groups. The main reason is $\mathcal{G}_{hard}(t)$s are quite different among iterations, keep switching entire training sets will lead to high oscillations, which slows the training convergence.

\subsection{The Representation Learning Algorithm}\label{subsec:algorithm}

The entire representation learning framework is shown in Figure~\ref{fig:architecture}. Specifically, given a set of limited training set $\mathcal{X}$ and a validation set $\mathcal{V}$ with corresponding crowdsourced labels separately, our proposed framework first generates sufficient training groups by pairing both positive and negative examples. Then, all training groups will be fed into a DNN which maximizes the conditional likelihood of retrieving the positive examples in Eq.(\ref{equ:weight_object}). 
For making the training process to be more efficient, in each training iteration $t$, we select hard training groups $\mathcal{G}(t)$ and use them to optimize the model parameters $\mathbf{\Omega}(t)$. 
Formally, we summarize the entire procedure of our proposed framework for learning effective representations from limited data with crowdsourced labels via adaptive hard example selection in Algorithm~\ref{alg:algorithm1}. 

\begin{algorithm}
\caption{Representation learning from limited crowdsourced labels via hard example selection.}
\label{alg:algorithm1}
\KwIn{training data $\mathcal{X}$, validation data $\mathcal{V}$} 
\KwOut{learned parameters $\hat{\bm{\Omega}}$}
initialize model parameters $\bm{\Omega}(0)$\;
select base training groups $\mathcal{G}_{base}$ randomly\;
\For{$t: 1 \rightarrow T$}{
    initialize $\mathcal{X}_{hard}(t) = \emptyset$\;
    \eIf{$t = 1$}{
    initialize $\mathcal{G}_{hard}(t) = \emptyset$\;
    }
    {obtain mis-predicted example set $\mathcal{V}_{miss}(t - 1)$ from $\mathcal{V}$ based on $\bm{\Omega}(t - 1)$\;
        \For{$\mathbf{v}'_j \in \mathcal{V}_{miss}(t - 1)$}{
            find the most similar $\mathbf{x}'_{j}$ to $\mathbf{v}'_j$ from $\mathcal{X}$\;
            update $\mathcal{X}_{hard}(t - 1) = \mathcal{X}_{hard}(t - 1) \cup \mathbf{x}'_{j}$\;
        }
        create $\mathcal{G}_{hard}(t - 1)$ from $\mathcal{X}_{hard}(t - 1)$\;
    }
    get $\mathcal{G}(t) = \left\{\mathcal{G}_{base}, \mathcal{G}_{hard}(t - 1) \right\}$\;
    conduct model learning to obtain $\bm{\Omega}(t)$ using $\mathcal{G}(t)$\;
    }
\end{algorithm}

In addition, when training the proposed model, we adopt several strategies for training a better model. First, at the beginning stage of model training, 5 iterations of warm-up training are performed on $\mathcal{G}_{base}$ before applying hard example mining process, which ensures a good model initialization.
Second, for stable performance in each training iteration, if the prediction accuracy of the current model on the validation data set $\mathcal{V}$ is less than $70\%$, we will temporarily abort the hard example selection process in this iteration and use the initial training groups $\mathcal{G}_{base}$ as the replacement to train the proposed model. 
\section{Experiments}\label{sec:experiments}

In this section, we evaluate our proposed representation learning framework on three real educational data sets. First, we provide a brief introduction about data collection process of these three data sets and discuss the way to extract raw features from collected raw data. Then, we conduct qualitative analysis about the representations learned by our proposed framework and a wide range of baselines through a binary classification task. Next, we visualize the representations learned by our proposed framework and investigate the performance of different distance functions used in our adaptive hard example selection strategy. Lastly, we demonstrate the online performance of our proposed framework achieved in a real-world educational practice.

\subsection{Experimental Settings}

\subsubsection{Data Sets}
To comprehensively evaluate our proposed framework, we conduct a series of experiments on three real-world data sets collected from a third-party educational platform\footnote{https://www.speiyou.com}.

\begin{itemize}
    \item \textbf{fluency-1\&2}. We collect 743 oral assignment submissions on math questions from students in 1st \& 2nd grades. The oral assignments (in audio format) ask students to talk about their mental thinking processes of how to solve math questions. We invite 5 annotators to assess whether the students' entire speeches are fluent and assign score 0 or 1 to each audio assignment. 
    \item \textbf{fluency-4\&5}. Similar to \textbf{fluency-1\&2}, we collect 1965 audio assignments in 4th \& 5th grades. Since students in 4th \& 5th grades are in the adolescent stage and their vocal cords fluctuate a lot compared to students in 1st \& 2nd grades, we explicitly separate them from \textbf{fluency-1\&2} data set. Each of the audio submission is graded by 11 annotators. 
    \item \textbf{preschool}. We collect 1767 speeches from a preschool presentation competition. The task is to evaluate the quality of the students' presentation speeches. Each presentation recording is rated by 11 annotators by giving a binary label, i.e., pass or fail.
\end{itemize}

We randomly split each data set into three parts, i.e., training set, validation set and test set, separately. For test set, we invite teaching experts to annotate each audio and the expert labels (0 or 1) are considered as the ground truth for the evaluation purpose. Since all these three data sets are consisted of audio clips, which cannot be used by our proposed model as well as existing related models directly, we need to extract raw features from each data set to construct three applicable data sets for experiments using.

\subsubsection{Raw Feature Extraction}\label{subsubsec:feature}

We briefly explain our raw feature extraction process. In this work, we extract both prosodic features and linguistic features from each audio clip. Prosodic features, such as signal energy, loudness, mel-frequency cepstral coefficients (MFCC), etc., are obtained by applying some automatic audio processing toolkits such as OpenSMILE\footnote{https://www.audeering.com/opensmile/}. To obtain linguistic features, such as statistics of part-of-speech tags, number of consecutive duplicated, number of interregnum words, etc., we feed audio clips into an automatic speech recognition (ASR) model to obtain structured text data. Please note that the ASR not only generates the text transcriptions but also the start and end timestamps for each sentence, which is very useful for computing the important features such as voice speed, silence duration percentage, etc. In summary, raw features extracted from the aforementioned raw data can be classified into four categories:

\begin{enumerate}
    \item \textbf{Word level features}, which contain features such as statistics of part-of-speech tags, number of consecutive duplicated words, number of interregnum words\footnote{Interregnum word is an optional part of a disfluent structure that consists of a filled pause \emph{uh} or a discourse marker \emph{I mean}.}, etc.
    \item \textbf{Sentence level features}, which contain different statistics from sentence perspective, such as distribution of clip vice length of each sentence, number of characters in each sentence, voice speed of each sentence, number of characters in each sentence, voice speed of each sentence, etc.
    \item \textbf{Instance (clip) level features}, which contain features like total number of characters and sentences, number of long silence that is longer than 5 seconds, the proportion of effective talking time to clip duration, etc.
    \item \textbf{Prosodic features}, which contain speech-related features such as signal energy, loudness, MFCC, etc.
\end{enumerate}

After extracting raw features from all of raw data sets separately, we can construct three corresponding applicable data sets for experiments using. The statistical information of these three new constructed data sets is listed in Table~\ref{tab:Statistics}. Here features contained in both fluency-1\&2 and fluency-4\&5 data sets are all linguistic features and in preschool data set are all prosodic features. Moreover, positive ratio measures the ratio between the number of positive examples and total examples in the training set.
\begin{table}[ht]
    \centering
    \caption{Statistics of data sets used in experiments.}
    \begin{tabular}{c|c|c|c}
    \hline
        & fluency-1\&2 & fluency-4\&5 & preschool \\
        \hline
        \hline
        \# training & 453 & 1,377 & 1,236 \\
        \# validation & 113 & 293 & 265 \\
        \# test & 177 & 295 & 266 \\
        \# annotator & 5 & 11 & 11 \\
        \# feature & 50 & 50 & 1,582 \\
        positive ratio & 0.689 & 0.837 & 0.654 \\
        \hline
    \end{tabular}
    \label{tab:Statistics}
\end{table}

\subsubsection{Baselines}
In order to assess the effectiveness of our proposed framework on learning effective representations from limited crowdsourced data, we select a series of models ranging from classic machine learning models to advanced deep learning models as baselines for comparing performance on a classification task. The selected baselines can be roughly categorized into three different groups.

1) Group 1: \emph{Popular Classification Methods}. The first group includes several popular classification methods that work on the raw features space directly. Since all these methods are designed for data with clean labels, we use the majority voting from the crowdsourced labels to infer the true labels for these methods.
\begin{itemize}
    \item \textbf{LR}, i.e., linear regression, which is a fundamental and commonly used linear model for discovering the relationship between variables or predicting possible values for some variables.
    \item \textbf{GBDT} i.e., gradient boosted decision trees~\cite{friedman2001greedy}, which is an classification model that utilizes an ensemble of decision trees to predict target labels.
    \item \textbf{SVM}, i.e., support vector machine, which is the most famous supervised learning model for non-linear classification tasks through the kernel trick.
\end{itemize}
    
2) Group 2: \emph{Representation Learning with Limited Labels}. The second group contains some effective representation learning methods specified for limited labels. Similarly, the majority voting is also introduced for these methods to infer the true labels.
\begin{itemize}
    \item \textbf{SiameseNet}, which is a Siamese neural network that to be trained on pairs of examples for learning representations~\cite{koch2015siamese}, so that the distance between a pair of examples is minimized if they're from the same class and is greater than some margin value if they represent different classes.
    \item \textbf{FaceNet}, which is a Triplet neural network that aims to learn representations~\cite{schroff2015facenet} such that the anchor is closer to the positive examples than it is to the negative examples by some margin value.
    \item \textbf{RelationNet}, which is a relation neural network for handling few-shot learning problem~\cite{sung2018learning} through learning a deep distance metric to compare a small number of images within episodes.
\end{itemize}

3) Group 3: \emph{Representation Learning with Limited Crowdsourced Labels}. The third group covers methods that can directly learn effective representations from limited crowdsourced labels without pre-processing crowdsourced labels.
\begin{itemize}
    \item \textbf{RLL-Bayesian}, which is an effective representation learning model~\cite{xu2019learning} trained on all training groups to learn embeddings from limited crowdsourced labeled data with confidence score estimated by a Bayesian approach.
\end{itemize}

We name our proposed framework as \textbf{RECLE}, which indicates the primary goal of our work in this paper, i.e., \underline{re}presentation learning with \underline{c}rowdsourced \underline{l}abels via adaptive hard \underline{e}xample selection. Our RECLE framework also belongs to the Group 3 since it targets to solve the same problem with the RLL-Bayesian model. We use \emph{TensorFlow}\footnote{https://www.tensorflow.org/} to implement all methods in Group 2 and Group 3 and methods in Group 1 are implemented by \emph{scikit-learn}\footnote{https://scikit-learn.org/} library. To encourage reproducible results, we make our code available online at \url{https://github.com/tal-ai/RECLE}.

\subsection{Experimental Results}

\subsubsection{Representation Learning Prediction Performance}

In order to verify whether our proposed framework could learn more useful and meaningful representations than all baselines, we introduce a linear regression model to perform a binary classification task trained on the representations learned by methods in Group 2 and Group 3 separately. For other three baselines, i.e., LR, GBDT and SVM, we train them on the raw features directly. Here we report the best prediction performance with respect to five different evaluation metrics of each model on three aforementioned real data sets in Tables \ref{tab:class_result_flu-12} - \ref{tab:class_result_preschool}.

As shown in these three tables, our RECLE framework achieves best prediction performance on all three real data sets under most evaluation metrics.
We think three main factors result in this phenomenon.
First, since the designed grouping strategy is able to alleviate the insufficient training example problem and provide much more representative data examples to train models, methods with the grouping-based strategy (methods in Group 2 and Group 3) produce significantly better results compared to methods in Group 1 that trained on the raw feature space with limited training examples.   
Second, because all methods in Group 1 and Group 2 need to infer the true labels from the crowdsourced labels in advance by majority voting, some noises are inevitably introduced into training data and some important information may be lost after this pre-processing stage. In contrast, both RLL-Bayesian and RECLE can learn effective representations from crowdsourced labels directly with the help of the Bayesian confidence estimation mechanism introduced in Section~\ref{subsec:bayesian}.
Third, compared to RLL-Bayesian proposed in~\cite{xu2019learning}, which trained on all training groups and does not consider mining hard examples during training process, the adaptive hard example selection procedure adopted in RECLE makes amendments to mistakes in the previous iteration, which can generate more robust representations. 
In addition, all models perform better on the fluency-1\&2 and fluency-4\&5 data sets than on the preschool data set. We believe the reason is that the prosodic features contained in the preschool data set cannot provide deterministic and dominant information to identify the fluency level in each audio clip as much as the linguistic features covered in the both fluency-1\&2 and fluency-4\&5 data sets.
\begin{table}[ht]
    \centering
    \caption{Prediction results on the fluency-1\&2 data set.}
    \begin{tabular}{c|c|c|c|c|c}
    \hline
        Model & Accuracy & Precision & Recall & F1 score & AUC \\
        \hline
        \hline
        LR & 0.825 & 0.842 & 0.918 & 0.878 & 0.908 \\
        GBDT & 0.819 & 0.888 & 0.844 & 0.866 & 0.888 \\
        SVM & 0.701 & 0.697 & \bf{1.0} & 0.822 & 0.708 \\
        \hline
        SiameseNet & 0.881 & 0.917 & 0.910 & 0.914 & 0.892 \\
        FaceNet & 0.870 & \bf{0.930} & 0.877 & 0.903 & 0.917 \\
        RelationNet & 0.876 & 0.917 & 0.902 & 0.909 & 0.926 \\
        \hline
        RLL-Bayesian & 0.836 & 0.904 & 0.852 & 0.878 & 0.919 \\
        RECLE & \bf{0.887} & 0.925 & 0.910 & \bf{0.917} & \bf{0.926} \\
        \hline
    \end{tabular}
    \label{tab:class_result_flu-12}
\end{table}

\begin{table}[ht]
    \centering
    \caption{Prediction results on the fluency-4\&5 data set.}
    \begin{tabular}{c|c|c|c|c|c}
    \hline
        Model & Accuracy & Precision & Recall & F1 score & AUC \\
        \hline
        \hline
         LR & 0.844 & 0.855 & 0.980 & 0.913 & 0.732 \\
         GBDT & 0.827 & 0.855 & 0.955 & 0.902 & 0.752 \\
         SVM & 0.837 & 0.837 & \bf{1.0} & 0.911 & 0.548 \\
         \hline
         SiameseNet & 0.848 & 0.846	& \bf{1.0} & 0.917 & 0.706 \\
         FaceNet & 0.851 & 0.849 & \bf{1.0} & 0.918 & 0.736 \\
         RelationNet & 0.848 & 0.851 & 0.992 & 0.916 & 0.718 \\
         \hline
         RLL-Bayesian & 0.851 & \bf{0.875} & 0.960 & 0.915 & \bf{0.771} \\
         RECLE & \bf{0.854} & 0.862 & 0.984 & \bf{0.919} & 0.749 \\
         \hline
    \end{tabular}
    \label{tab:class_result_flu-45}
\end{table}

\begin{table}[ht]
    \centering
    \caption{Prediction results on the preschool data set.}
    \begin{tabular}{c|c|c|c|c|c}
    \hline
        Model & Accuracy & Precision & Recall & F1 score & AUC \\
        \hline
        \hline
         LR & 0.729 & 0.743 & 0.897 & 0.812 & 0.734 \\
         GBDT & 0.733 & 0.749 & 0.891 & 0.814 & 0.763 \\
         SVM & 0.654 & 0.654 & \bf{1.0} & 0.791 & 0.492 \\
         \hline
         SiameseNet & 0.707 & 0.698 & 0.971 & 0.813 & 0.669 \\
         FaceNet & 0.718 & 0.712 & 0.954 & 0.816 & 0.664 \\
         RelationNet & 0.737 & 0.734 & 0.937 & 0.823 & 0.701 \\
         \hline
         RLL-Bayesian & 0.752 & \bf{0.784} & 0.856 & 0.819 & 0.767 \\
         RECLE & \bf{0.774} & 0.777 & 0.920 & \bf{0.842} & \bf{0.782} \\
         \hline
    \end{tabular}
    \label{tab:class_result_preschool}
\end{table}

\subsubsection{Evaluation of Learned Representations}

\begin{figure}[ht]
\centering
\subfigure[Raw feature space.]{
\begin{minipage}[ht]{0.47\linewidth}
\centering
\includegraphics[width=1.7in]{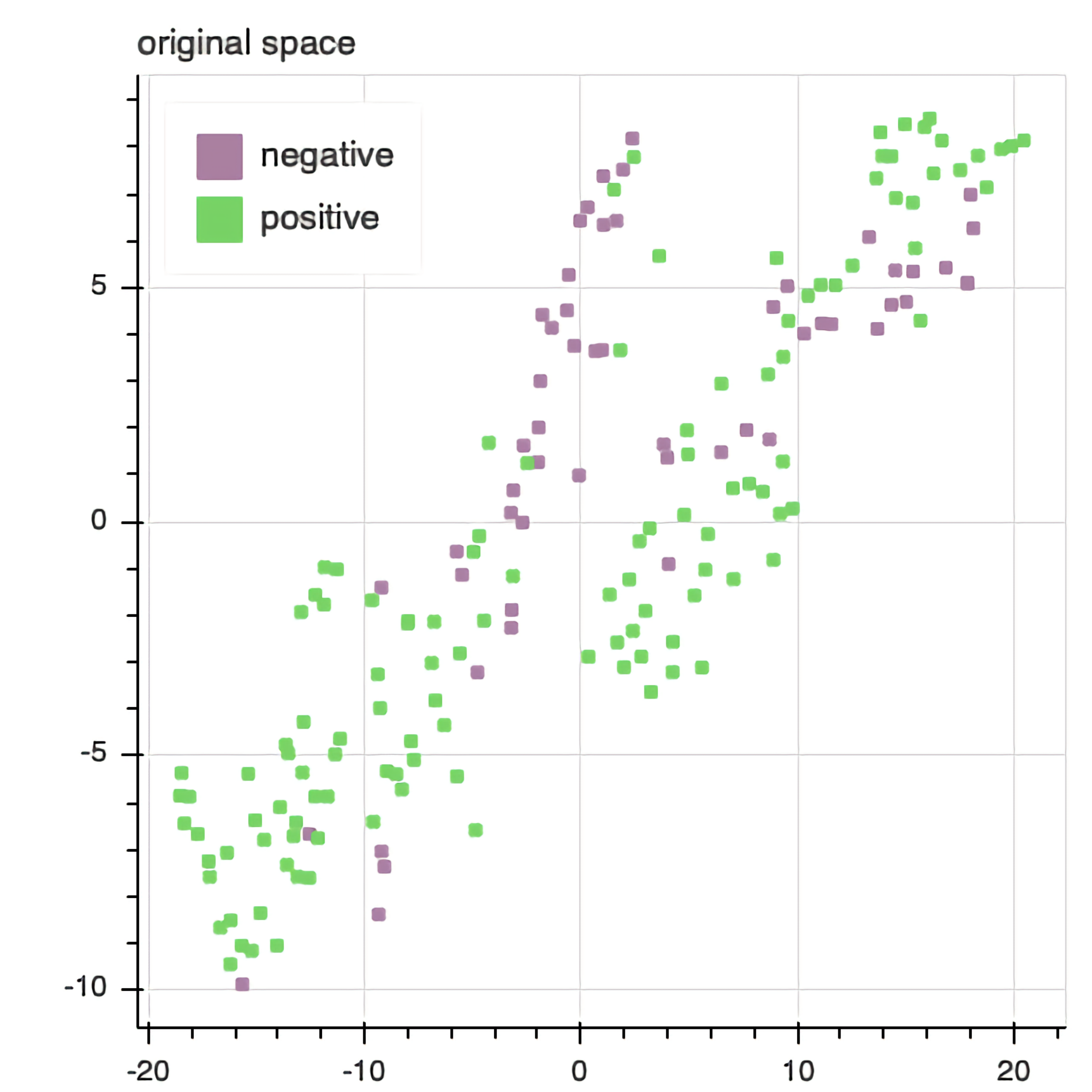}
\end{minipage}%
}%
\hspace{0.15cm}
\subfigure[Learned embedding space.]{
\begin{minipage}[ht]{0.47\linewidth}
\centering
\includegraphics[width=1.7in]{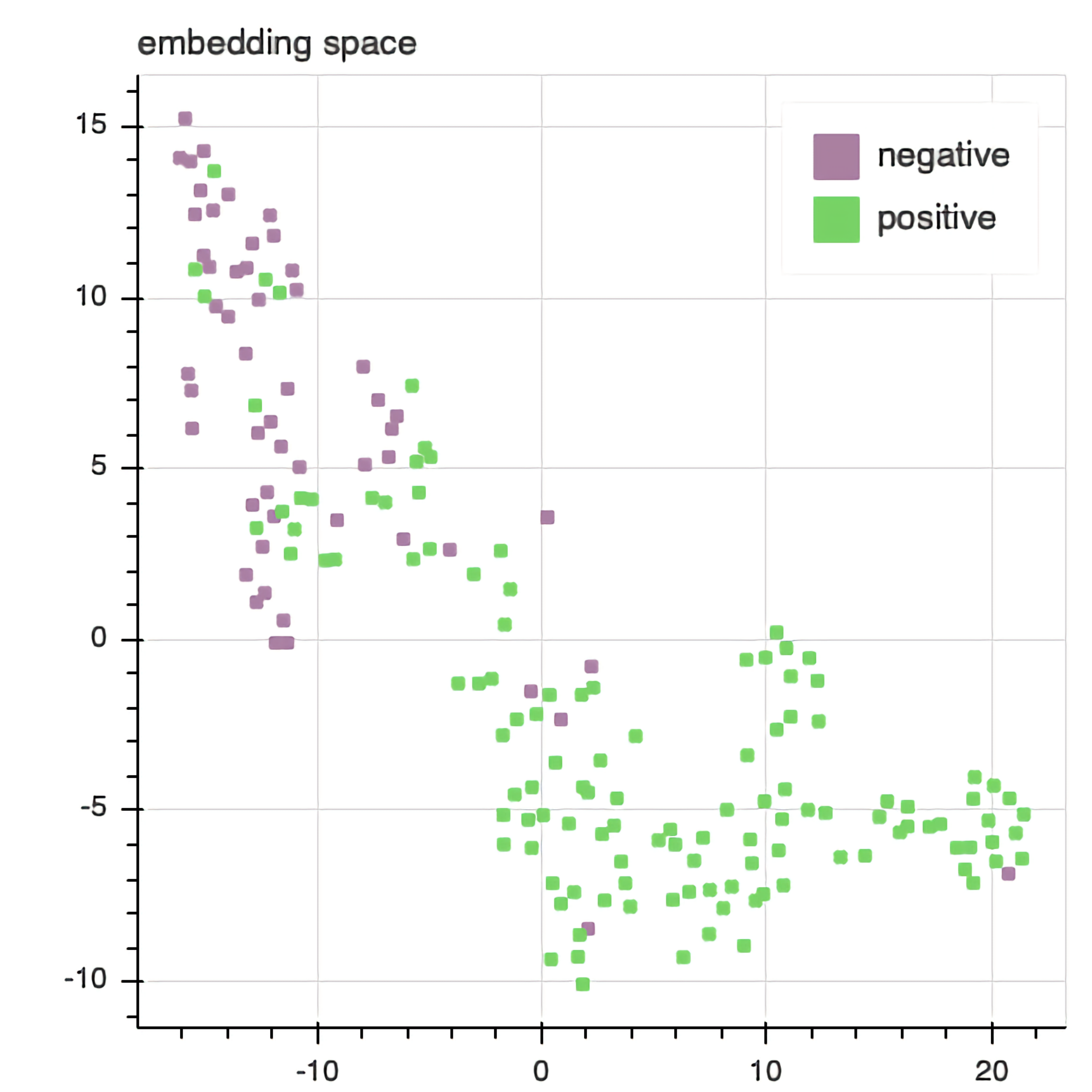}
\end{minipage}
}%
\centering
\caption{Visualization of raw data and learned representations on the fluency-1\&2 data set.}
\label{fig:visual_flu-12}
\end{figure}

\begin{figure}[ht]
\centering
\subfigure[Raw feature space.]{
\begin{minipage}[ht]{0.47\linewidth}
\centering
\includegraphics[width=1.7in]{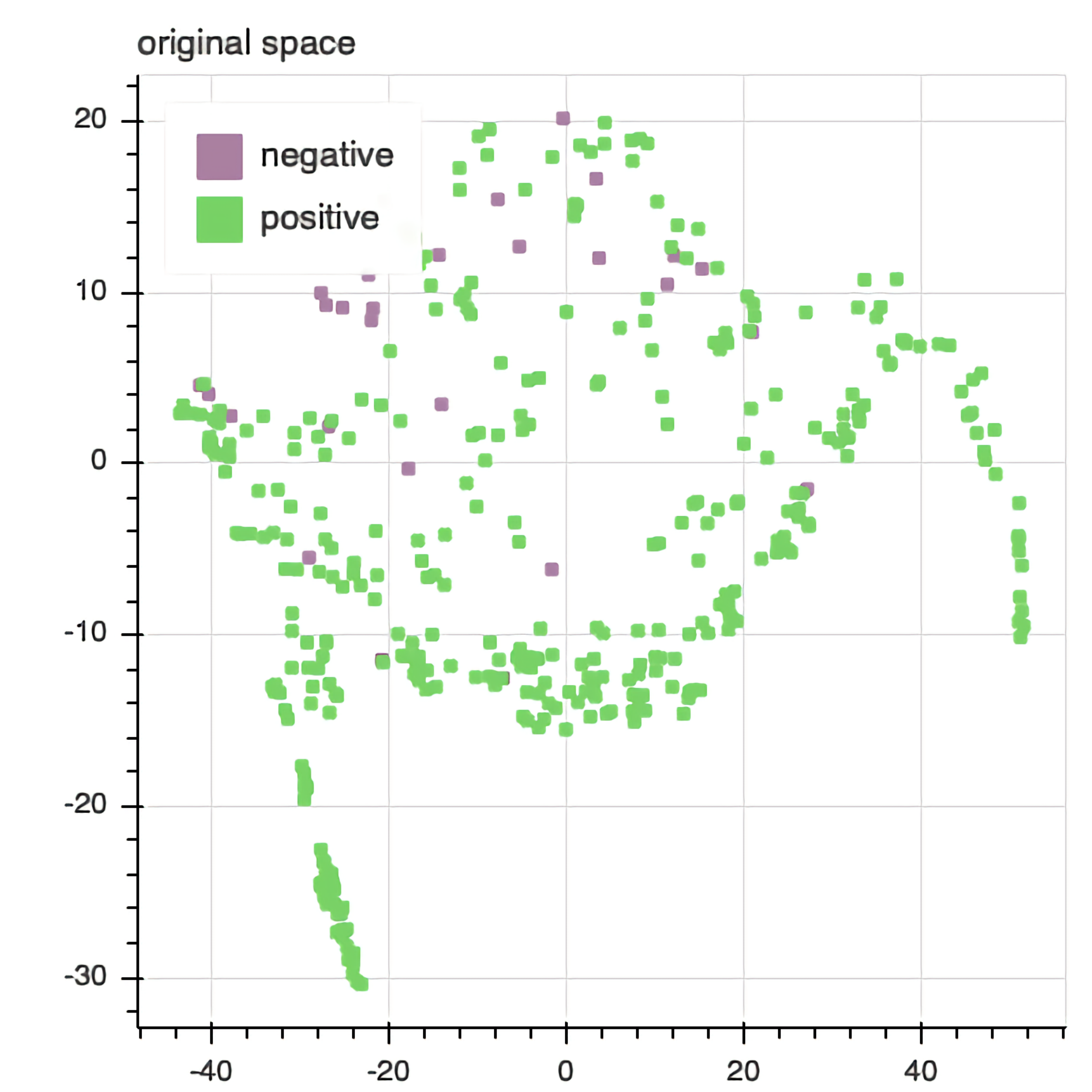}
\end{minipage}%
}%
\hspace{0.15cm}
\subfigure[Learned embedding space.]{
\begin{minipage}[ht]{0.47\linewidth}
\centering
\includegraphics[width=1.7in]{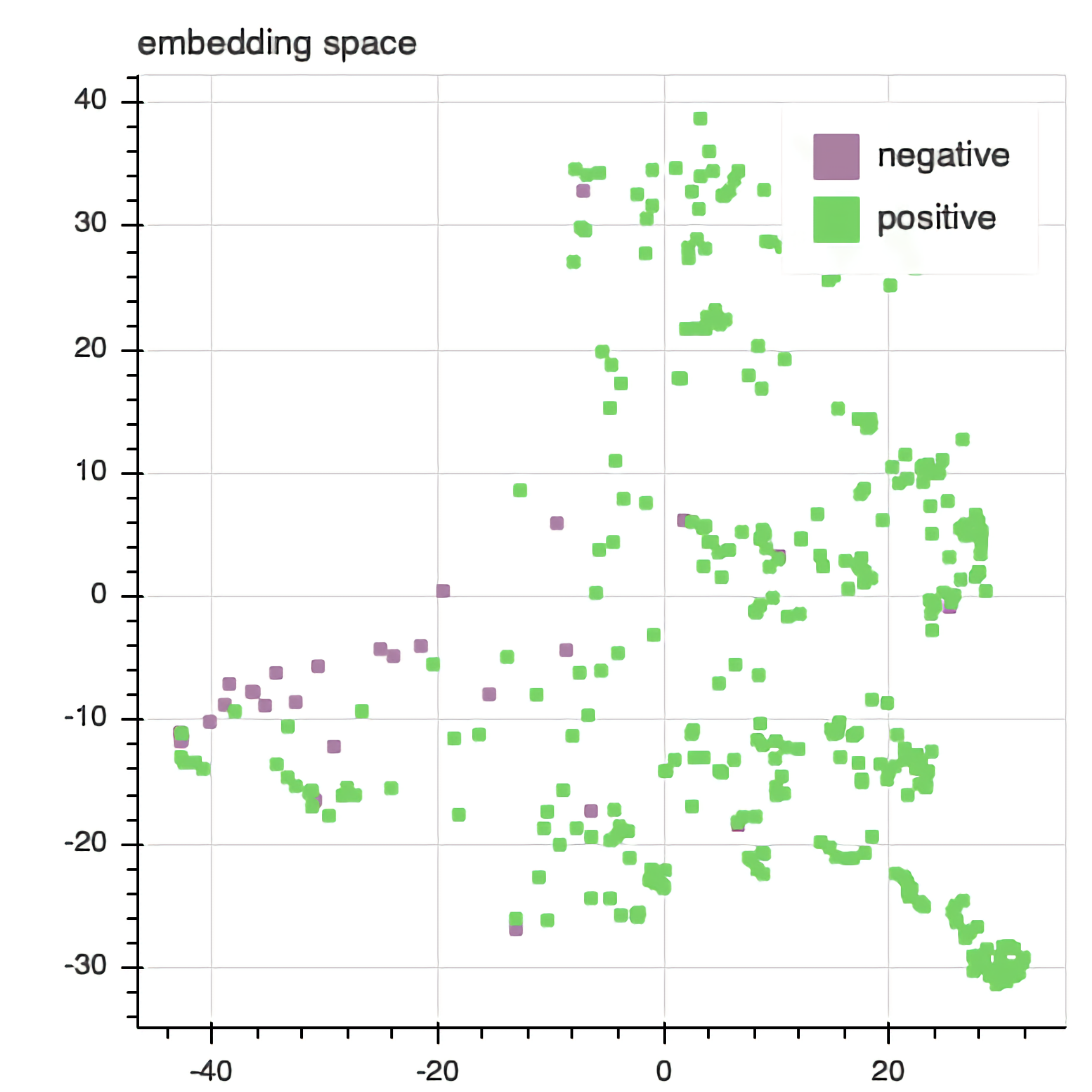}
\end{minipage}
}%
\centering
\caption{Visualization of raw data and learned representations on the fluency-4\&5 data set.}
\label{fig:visual_flu-45}
\end{figure}

\begin{figure}[ht]
\centering
\subfigure[Raw feature space.]{
\begin{minipage}[ht]{0.47\linewidth}
\centering
\includegraphics[width=1.7in]{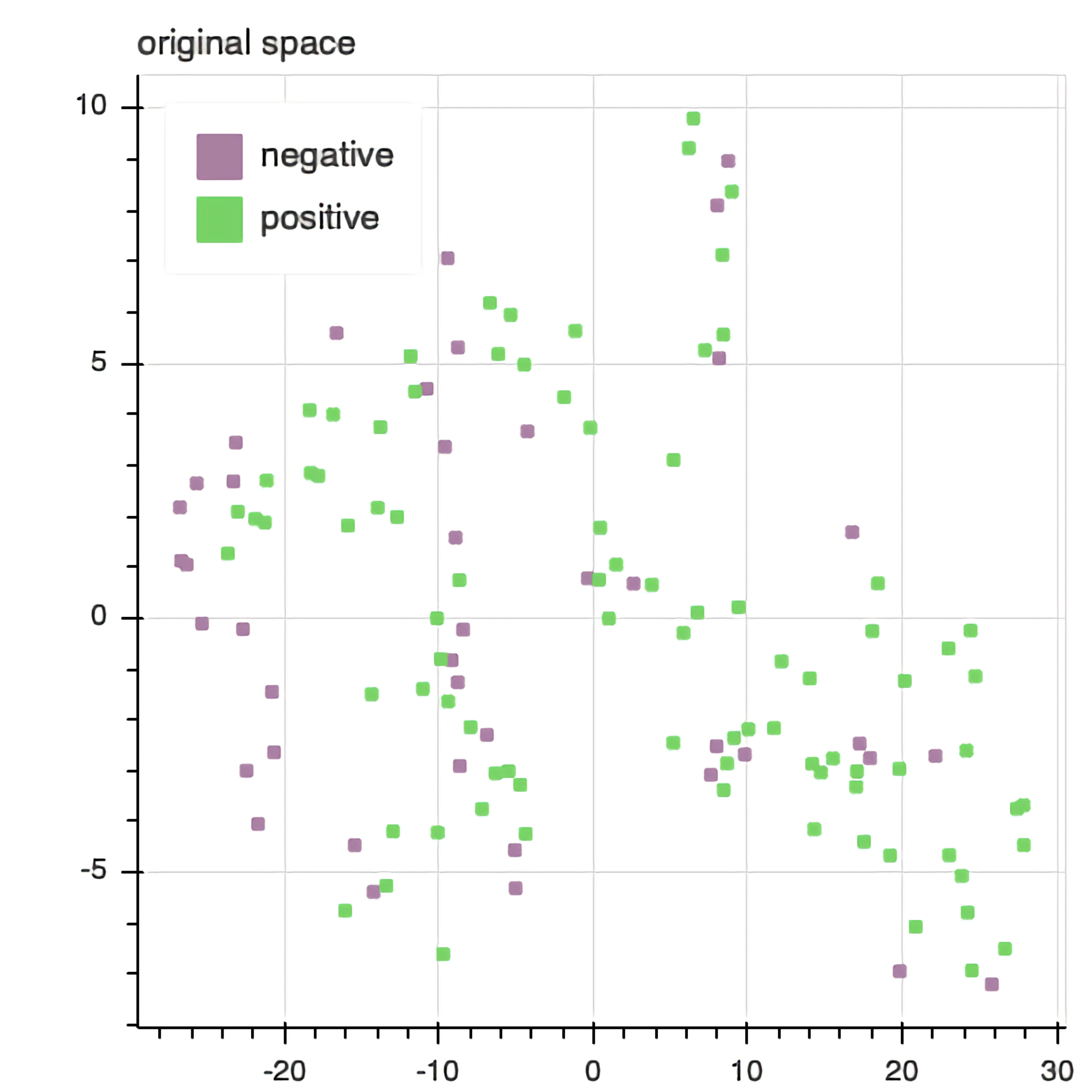}
\end{minipage}%
}%
\hspace{0.15cm}
\subfigure[Learned embedding space.]{
\begin{minipage}[ht]{0.47\linewidth}
\centering
\includegraphics[width=1.7in]{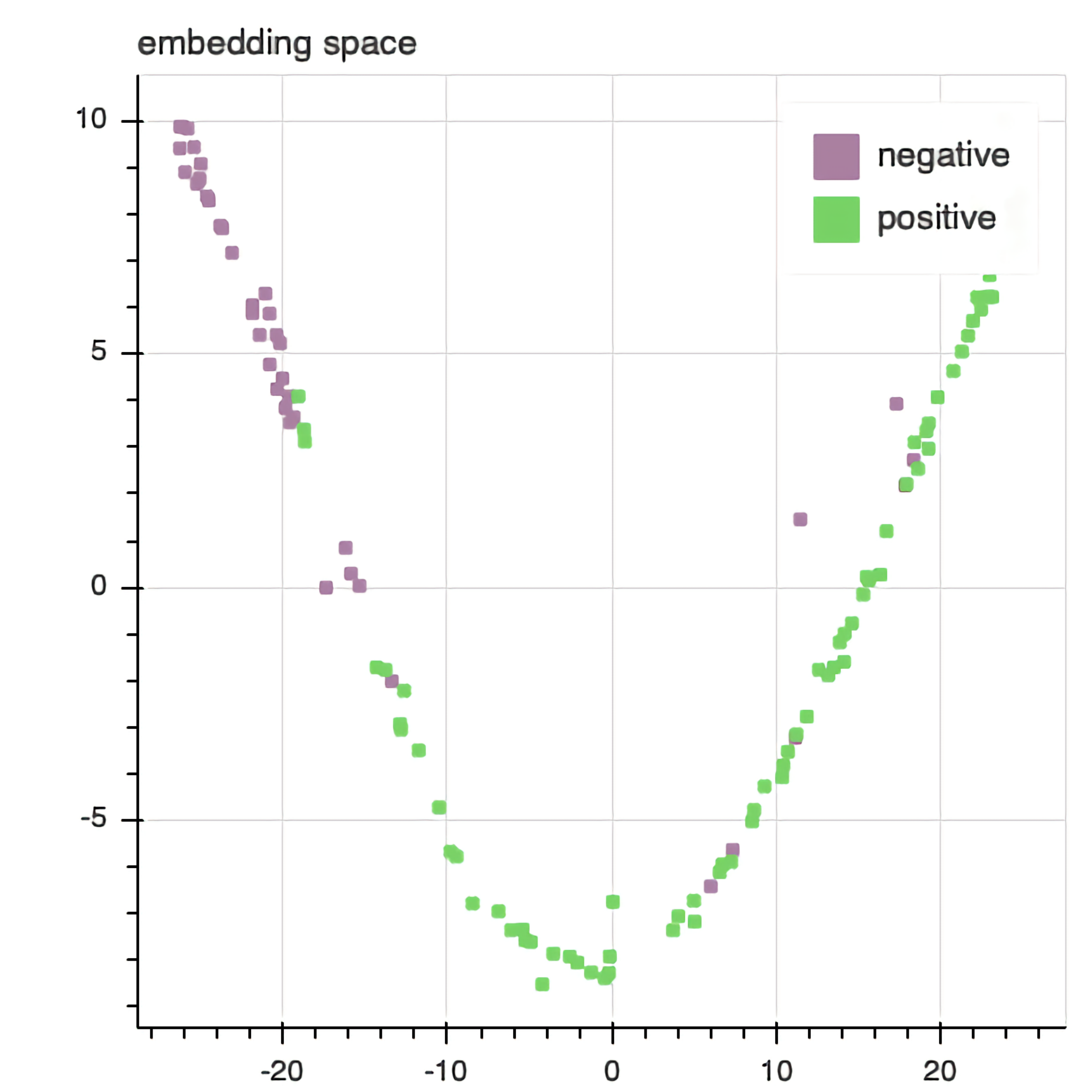}
\end{minipage}
}%
\centering
\caption{Visualization of raw data and learned representations on the preschool data set.}
\label{fig:visual_preschool}
\end{figure}

For providing a straightforward way to demonstrate the effectiveness of our proposed model, we visualize the representations learned by our RECLE framework in the embedding space as well as the raw data in the raw feature space for all three real data sets. Considering the raw data are high-dimensional data and the learned representations still have relatively large dimension, we utilize t-SNE~\cite{maaten2008visualizing} to visualize this kind of high-dimensional data by conducting a dimension reduction operation from the original high-dimensional space into a 2-dimensional space. Figure~\ref{fig:visual_flu-12} to Figure~\ref{fig:visual_preschool} show the visualization results on the fluency-1\&2, fluency-4\&5 and preschool data sets, respectively.  

Here we can see, comparing with the positive examples and negative examples that are highly interleaved with each other in the raw feature space, the learned representations of examples of these two classes in the embedding space can be separated from each other to some extent, which could greatly alleviate the difficulty of training better classification models for the binary classification task. Therefore, in other words, our proposed model is able to learn useful and meaningful representations from limited crowdsourced data and, hence, benefit the classification task.  

\subsubsection{Impact of Different Distance Metrics}\label{subsec:distance}
In order to fully understand the performance of our proposed framework, we empirically measure the effects of different distance metrics used when adaptively selecting hard examples. We choose the following three common approaches to select hard examples:
\begin{itemize}
    \item \textbf{Cosine}, i.e., cosine distance function.
    \item \textbf{L1}, i.e., Manhattan distance function.
    \item \textbf{L2}, i.e., Euclidean distance function.
\end{itemize}
For each real data set, we apply the same parameter settings of our RECLE framework and only vary distance metrics to study the effects of using different distance metrics. We repeat experiments 10 times for each distance metric and report the average prediction results obtained on the corresponding test data in Tables \ref{tab:distance_result_flu-12} - \ref{tab:distance_result_preschool}.

As we can see from these three tables, the prediction performance changes when we vary the adopted distance function. Overall, the cosine distance function outperforms other distance metrics on the fluency-1\&2 and fluency-4\&5 data sets which consist of linguistic features while the Euclidean distance function beats all other distance metrics on the preschool data set which only contains prosodic features. Based on that, we can adopt different distance functions in our RECLE framework, according to the different characteristics of the training data set.

\begin{table}[ht]
    \centering
    \caption{Prediction results with different distance metrics on the fluency-1\&2 data set.}
    \begin{tabular}{c|c|c|c|c|c}
    \hline
        Model & Accuracy & Precision & Recall & F1 score & AUC \\
        \hline
        \hline
        Cosine & \bf{0.852} & \bf{0.898} & \bf{0.887} & \bf{0.892} & \bf{0.904} \\
        L1 & 0.838 & 0.887 & 0.877 & 0.882 & 0.893 \\
        L2 & 0.832 & 0.885 & 0.869 & 0.877 & 0.895 \\
        \hline
    \end{tabular}
    \label{tab:distance_result_flu-12}
\end{table}

\begin{table}[ht]
    \centering
    \caption{Prediction results with different distance metrics on the fluency-4\&5 data set.}
    \begin{tabular}{c|c|c|c|c|c}
    \hline
        Model & Accuracy & Precision & Recall & F1 score & AUC \\
        \hline
        \hline
        Cosine & \bf{0.847} & 0.853 & \bf{0.987} & \bf{0.915} & 0.752 \\
        L1 & 0.840 & \bf{0.857} & 0.970 & 0.910 & \bf{0.757} \\
        L2 & 0.840 & 0.856 & 0.972 & 0.911 & 0.754	 \\
        \hline
    \end{tabular}
    \label{tab:distance_result_flu-45}
\end{table}

\begin{table}[ht]
    \centering
    \caption{Prediction results with different distance metrics on the preschool data set.}
    \begin{tabular}{c|c|c|c|c|c}
    \hline
        Model & Accuracy & Precision & Recall & F1 score & AUC \\
        \hline
        \hline
        Cosine & 0.753 & 0.766 & 0.900 & 0.827 & 0.753 \\
        L1 & 0.754 & 0.765 & 0.900 & 0.827 & 0.758 \\
        L2 & \bf{0.763} & \bf{0.772} & \bf{0.906} & \bf{0.833} & \bf{0.760} \\
        \hline
    \end{tabular}
    \label{tab:distance_result_preschool}
\end{table}

\subsubsection{Impact of Imbalanced Data Distribution}

\begin{table*}[ht]
    \centering
    \caption{Prediction results with different grouping strategies on the fluency-4\&5 data set.}
    \begin{tabular}{c|c|c|c|c|c}
    \hline
        Grouping strategy & Accuracy & Precision & Recall & F1 score & AUC  \\
        \hline
        \hline
        2 positive + $S-2$ negative & 0.840 & 0.857 & 0.970 & 0.910 & 0.757 \\
        2 negative + $S-2$ positive & 0.842 & 0.859 & 0.971 & 0.911 & 0.758 \\
        \hline
    \end{tabular}
    \label{tab:imbalanced}
\end{table*}

In Section~\ref{subsec:grouping}, we introduce our grouping strategy for alleviating the insufficient training data problem. Briefly, a new training grouping with size $S$ is consisted of 2 positive examples and $S - 2$ negative examples randomly picked up from a positive training set and a negative training set, respectively. However, since all three data sets used in experiments are collected from our educational practice, these three data sets cannot be naturally class balanced, as shown in Figure~\ref{fig:visual_flu-12} to Figure~\ref{fig:visual_preschool}. For studying whether the imbalanced data distribution affects the performance of our grouping strategy, we conduct a comparison experiment between two different grouping strategies on the fluency-4\&5 data set. The positive example ratio of fluency-4\&5 is 0.837, which means most of examples in this data set are positive examples. In addition to the grouping strategy adopted in our RECLE framework, we also design a new grouping strategy that uses 2 negative examples and $S - 2$ positive examples to create a new training group and integrate it into RECLE to perform a binary classification task on the fluency-4\&5 data set. We conduct experiments 10 times and report average results in Table~\ref{tab:imbalanced}.

As shown in Table~\ref{tab:imbalanced}, when we fix all model parameters and only change the way of creating training groups, there is almost no difference of prediction performance between two grouping strategies. In other words, our presented grouping strategy and proposed RECLE framework can work well on imbalanced limited data sets.

\subsection{Production Deployment \& Online Performance}

We build a real-world production pipeline to utilize RECLE for better predictive performance on automatic grading of students' submissions. More specifically, we use the neural network trained by RECLE to automatically evaluate students' free talks and give spoken language proficiency assessment scores. For each newly submitted recording of free talk, we first extract its raw features as discussed in Section \ref{subsubsec:feature}. Then we pass these raw features to the deployed neural network trained by RECLE to obtain its effective latent representations. After that, we conduct standard binary classification on the learned effective embeddings and output the final grading scores. 

We conduct the online A/B experiments to fully demonstrate the effectiveness of our RECLE framework. We incorporate RECLE into our spoken language proficiency assessment system\footnote{https://ai.100tal.com/dolphin} to predict language fluency from students' oral language skill exercises. By comparing to previous version that does not include RECLE framework for two weeks, we found that RECLE is able to achieve 3\%-5\% performance increase in terms of teachers' acceptance rate. The acceptance rate is defined as the number of model score accepted by professional teachers divided by the total number of predictions and its score between 1 to 5.

\section{Conclusion and Future Work}\label{sec:conclusion}

In this paper, we investigate the problem of leaning representations from limited data with crowdsourced labels. In order to deal with the scarcity and inconsistency issues brought by the small number of data examples annotated by crowd workers, we present a novel representation learning framework which can (1) automatically create a massive number of sample groups for the DNN training; (2) explicitly capture the inconsistency of crowdsourced labels and integrate it into model training process; and (3) adaptively select hard training groups to make the training process more efficient and sufficient. For verifying the effectiveness of our proposed framework RECLE, we compare RECLE with a wide range of baselines on three educational data sets. The extensive experimental results demonstrate that comparing with traditional discriminative representation learning methods, our RECLE framework is able to address the limited and inconsistent label problems simultaneously and learn effective embeddings from very limited data. 

In the future, we plan to extend this work from three aspects. First, we plan to incorporate information about individual crowd worker into the model training to reduce label noises and train more robust model. Second, we will study more advanced strategies of combining training groups at different training stages to improve the overall prediction accuracy further. Third, we are interested in experimenting with our proposed framework on other types of limited data with crowdsourced labels, such as cancer diagnosis in biomedical imaging.

\section*{Acknowledgments}
Wentao Wang and Jiliang Tang are supported by the National Science Foundation (NSF) under grant numbers IIS1907704, IIS1714741, IIS1715940 and IIS1845081. Guoliang Li is supported by NSF of China (61925205, 61632016). Zitao Liu is supported by Beijing Nova Program (Z201100006820068) from Beijing Municipal Science \& Technology Commission.


\ifCLASSOPTIONcaptionsoff
  \newpage
\fi



\bibliographystyle{IEEEtran}
\bibliography{reference}
%



%
\vspace{-0.3in}
\begin{IEEEbiography}[{\includegraphics[width=1in,height=1.25in,clip,keepaspectratio]{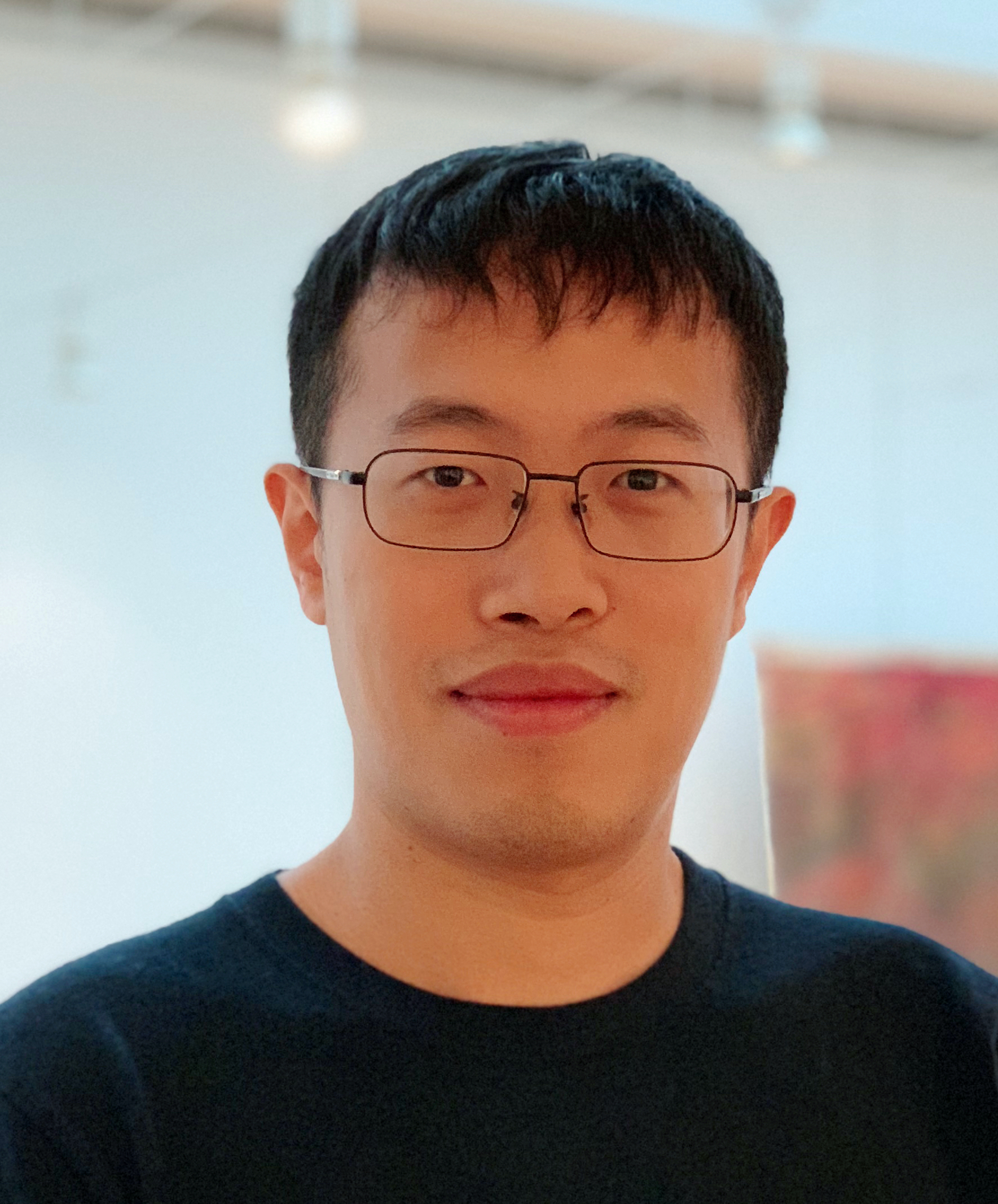}}]{Wentao Wang}
is a Ph.D. student in the computer science and engineering department at Michigan State University, supervised by Dr. Jiliang Tang. His research interests mainly lie in machine learning and data mining with a focus on learning from insufficient data. Before that, He received his bachelor's degree in Computer Science and Technology from Sichuan University in China.
\end{IEEEbiography}
\vspace{-0.43in}
\begin{IEEEbiography}[{\includegraphics[width=1in,height=1.25in,clip,keepaspectratio]{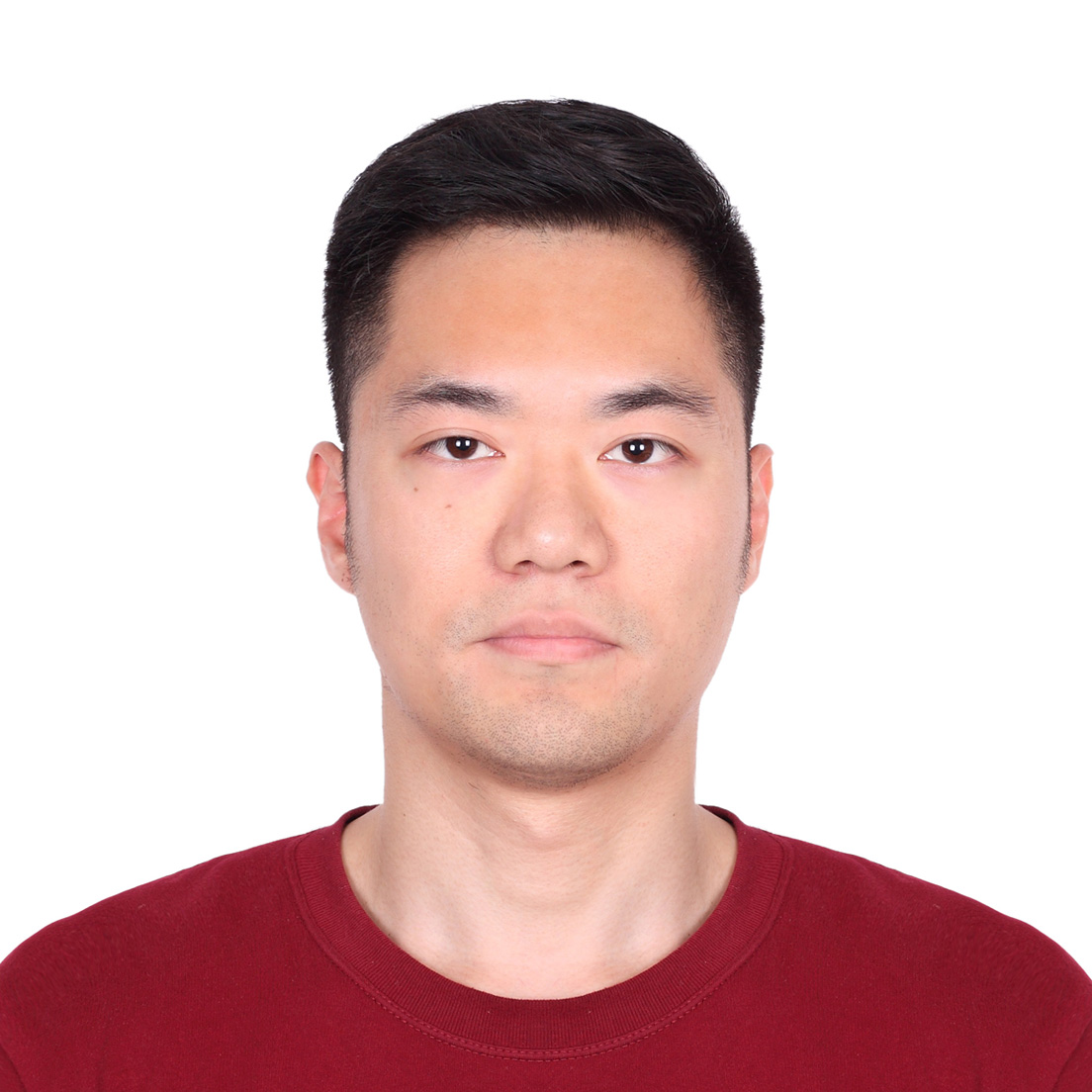}}]{Guowei Xu}
is a senior machine learning engineer at TAL Education Group (NYSE:TAL). His research focuses on data mining, natural language processing and its applications in education domains. He has built several influential systems in education field, such as AI-based Chinese spoken language proficiency assessment system, Chinese-English neural translation system, etc. Before joining TAL, he received his master degree in Electrical Engineering from Columbia University and his bachelor degree in Information Science from Beihang University.
\end{IEEEbiography}
\begin{IEEEbiography}[{\includegraphics[width=1in,height=1.25in,clip,keepaspectratio]{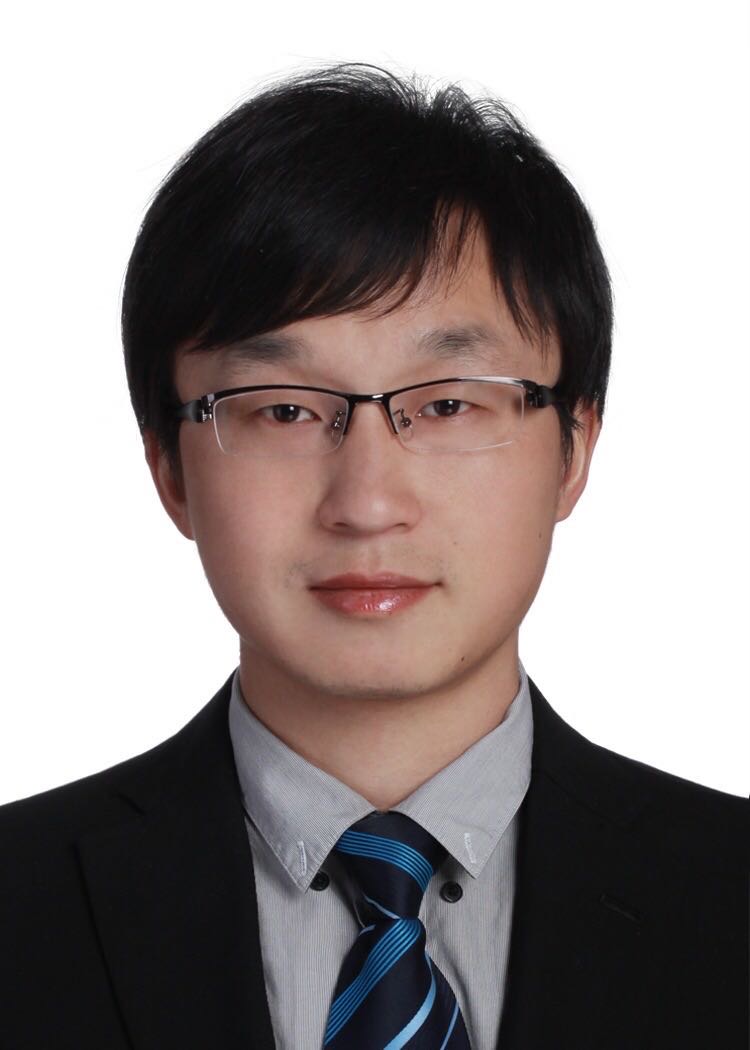}}]{Wenbiao Ding}
is the Head of Data Mining at TAL Education Group (NYSE:TAL). He has published several papers at top conference proceedings, such as ICDE, WWW, AIED, etc. He received his master's degree in computer science from the University of Science and Technology of China. Before joining TAL, Wenbiao was a senior research engineer at Sogou (NYSE:SOGO). He worked on information retrieval, large-scale data mining, natural language processing and their applications in search engine systems and recommendation systems. 
\end{IEEEbiography}
\vspace{-0.2in}
\begin{IEEEbiography}[{\includegraphics[width=1in,height=1.25in,clip,keepaspectratio]{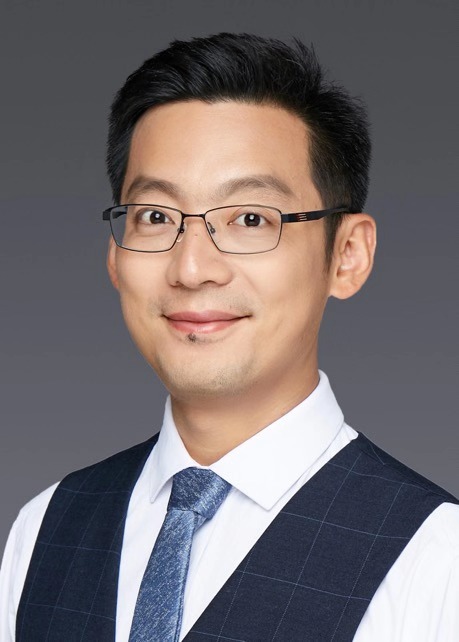}}]{Gale Yan Huang}
is the CTO \& BG President of the Open Education Platform Group (EPG) at TAL. He leads a team of thousands of engineers and researchers that focus on the in-depth application of frontier technology in the educational field. He is in charge of the EPG for numerous departments at TAL, including: Magic School, WISROOM, Weclassroom, First Leap and many more. These departments share advanced educational content and technical solutions that benefits key industry players to explore the joint force of science and education, and continuously promote the innovation and development of the education industry. Gale was the former Chief Architect of Baidu, the Director at the Tencent Research Institute, and Co-founder/Software Architect of PPLive (now PPTV).
\end{IEEEbiography}
\vspace{-0.2in}
\begin{IEEEbiography}[{\includegraphics[width=1in,height=1.25in,clip,keepaspectratio]{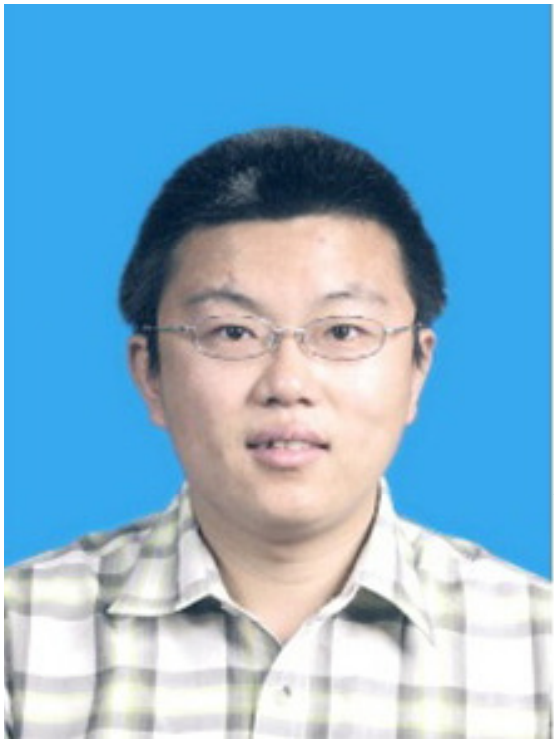}}]{Guoliang Li}
is currently working as a professor in the Department of Computer Science, Tsinghua University, Beijing, China. He received his PhD degree in Computer Science from Tsinghua University, Beijing, China in 2009. His research interests mainly include data cleaning and integration, crowdsourcing, and hybrid DB\&AI co-optimization.
\end{IEEEbiography}
\vspace{-0.2in}
\begin{IEEEbiography}[{\includegraphics[width=1in,height=1.25in,clip,keepaspectratio]{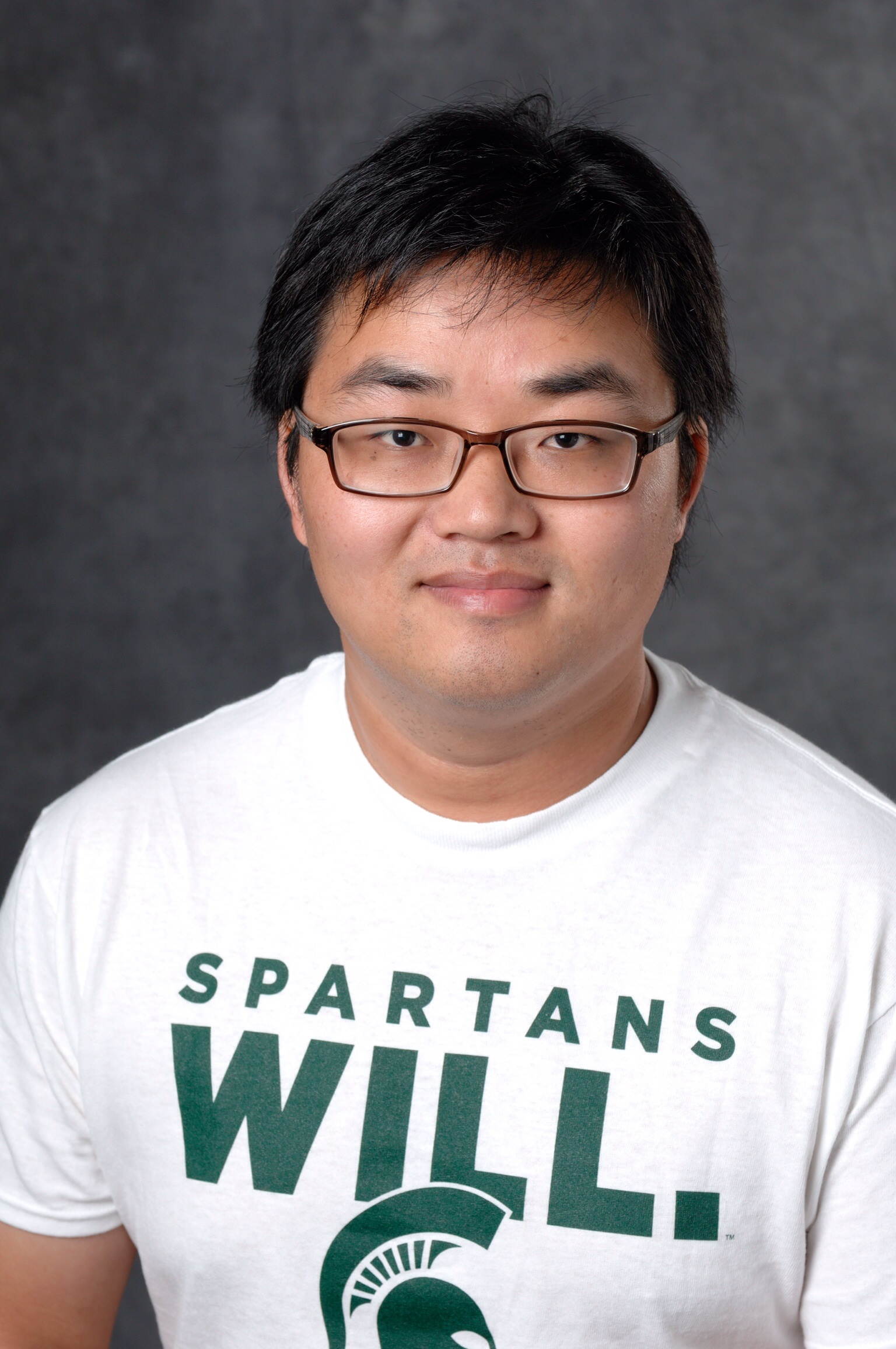}}]{Jiliang Tang}
is an assistant professor in the computer science and engineering department at Michigan State University since Fall 2016. Before that, he was a research scientist in Yahoo Research and got his PhD from Arizona State University in 2015. His research interests including social computing, data mining and machine learning and their applications in education. He was the recipients of 2019 NSF Career Award, the 2015 KDD Best Dissertation runner up and 6 best paper awards (or runner-ups) including WSDM2018 and KDD2016. He serves as conference organizers (e.g., KDD, WSDM and SDM) and journal editors (e.g., TKDD). He has published his research in highly ranked journals and top conference proceedings, which received thousands of citations and extensive media coverage.
\end{IEEEbiography}
\vspace{-0.2in}
\begin{IEEEbiography}[{\includegraphics[width=1in,height=1.25in,clip,keepaspectratio]{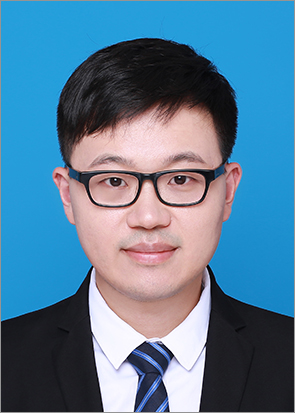}}]{Zitao Liu}
is the Director of AI Solution at TAL Education Group (NYSE:TAL). His research is in the area of machine learning, and includes contributions in the areas of artificial intelligence in education, multimodal knowledge representation and user modeling. He has published his research in highly ranked conference proceedings and serves as the executive committee of the International AI in Education Society and top tier AI conference/workshop organizers/program committees. Before joining TAL, Zitao was a senior research scientist at Pinterest and received his Ph.D degree in Computer Science from University of Pittsburgh.
\end{IEEEbiography}





\vfill


\end{document}